# From Minimal Existence to Human Definition: The CES-IMU-HSG Theoretical Framework

Kei Itoh*

## Abstract

This study presents an inter-universal mathematical-logical framework constructed upon the minimal axiom Cogito, ergo sum (CES), integrating the Intermediate Meta-Universe (IMU) and the Hierarchical State Grid (HSG). The CES defines existence as a reflexive correspondence --'to be' and 'to be sayable'-- and positions any formal system, including ZFC or HoTT, as an attachable extension atop this minimal structure. The IMU functions as a registry of axiomatic dependencies that connect heterogeneous theories, employing the Institution-theoretic framework to ensure coherent inter-theoretical linkages. The HSG concretizes these ideas through categorical construction, defined by three orthogonal axes: the state-depth axis, the mapping-hierarchy axis, and the temporal axis incorporating the principle of 'no future reference.' Through these, the identity of 'definition = state' is formally established as a categorical property. Extending this structure to biological systems, the neural system is implemented as a 0-3D complex of neuron-function fields on the HSG, while its categorical extensions via fiberization over the material base enable the parallel integration of multiple physiological universes-neural, endocrine, learning, genetic, and input/output systems-into a coherent adjoint ensemble. Within this framework, human behavior and cognition emerge as temporal compositions of inter-universal algorithms constrained by the material base. Finally, by contrasting human cognition, which relies on external CES, with machine existence, this study introduces the concept of internal CES, wherein a machine grounds its own logic upon the factuality of its operation. This internal self-axiomatization establishes a continuous bridge between philosophical ontology and engineering implementation, providing a new foundation for the autonomous and self-defining existence of artificial intelligence.

* Ehime University Graduate School of Science and Technology. The author conducted this research independently while affiliated with Ehime University and utilized the university's library and electronic resources. The research was conducted without specific financial support from external funding sources. The author declares no conflict of interest. While this English version constitutes the formal and citable edition of the work, the original manuscript was composed in Japanese, and some structural or rhetorical nuances may be more fully preserved in the original text.

## Introduction

This study is based on the inter-universal system theory established through the Hierarchical State Grid (HSG) and the Intermediate Meta-Universe (IMU), both introduced in Itoh (2025.08), and formalizes these frameworks within mathematical logic while applying them to the definition of "human" and the structural analysis of complex-system algorithms. In particular, the Cogito, ergo sum (CES) axiom is introduced as the minimal nucleus of the inter-universal axiom system. CES is defined as the minimal axiom consisting only of the two components "to exist" and "to be said to exist." In contrast, all other axioms are regarded as attachable auxiliary packs that can be layered upon it. The concept of existence is an inevitable syntactic prerequisite embedded in every act of definition, and therefore, this study takes the stance that "I think, therefore I am" should be externalized from its subjective proposition and positioned as a non-personal minimal syntactic axiom. Moreover, this study does not aim to ensure full formal rigor at the initial stage in the conventional sense, but rather to secure rigor as an "extensible skeleton" based on the design philosophy unique to this framework, leaving the detailed formal consistency and typing to subsequent studies. In addition, categorical constructions used in this study (adjunctions, monads, Kan extensions, skeletons/coskeletons, limits, colimits, etc.) follow the standard definitions in Mac Lane (1971), Riehl (2016), and Goerss–Jardine (1999), unless otherwise noted.

## Background and issues

In Itoh (2025.08), HSG was introduced as a notational system that places state depth, mapping hierarchy, and time on Cartesian coordinate axes, thereby allowing "definition = state" to be represented. IMU was introduced as a device for meta-defining a universe from the outside, under the premise of self-referential incompleteness (that we cannot fully describe ourselves). These ideas are naturally connected to inter-universal logical frameworks such as "change of universe (e.g., Lurie, 2009)" and IUT (Mochizuki, 2012). However, in previous studies, they remained at the level of notation, and their mathematical-logical foundations had not yet been established.

## Structure of the study

1. **Introduction of CES and the mathematical structure of IMU**
   In the inter-universal axiom systems, including IMU, it is considered that developing logic by attaching every other axiom "post hoc" upon the CES axiom (or another minimal axiom) as the only minimal base leads to a controlled theory construction. As mentioned above, the minimal configuration of IMU consists solely of CES, and all axiomatic objects, including notational forms, are added on top of it. For example, both ZFC and HoTT are positioned as different axiom packs within the same registry under the CES framework. IMU functions as a registry and dependency management system for such axiom packs, and when necessary, utilizes the institutional framework (signatures, translations, and satisfaction preservation) (e.g., Goguen and Burstall, 1992). However, such model-theoretic frameworks are regarded as optional tools useful for comparatively advanced logical constructions, such as interoperability among parallel universes, and are not mandatory requirements.
2. **Definition of HSG**
   HSG is introduced as one of the axioms within the framework of CES + IMU, within which various state objects are defined. First, by decomposing the structure of HSG from Itoh (2025.08), the minimal configuration of HSG is defined only by "the set of axes," "the index sets for each axis," and "the Cartesian product grid with tuple projection equalities (Cartesian notation)" (without any categorical structure). This allows for a freer and more general definition than the HSG in Itoh (2025.08). Such purely notational construction is referred to as a "notational axiom," which makes it possible to distinguish clearly between "axioms as mathematical structures" and "axioms as notational structures," which had previously been ambiguous. In this study, the numerical system HSG in Itoh (2025.08) is interpreted as the product 2-category $\prod_i \tau_{\leq 2}(C_i)$, where each axis is assigned a category C_i. The state-depth axis is interpreted as a "cumulative-axiom 2-category" equipped with adjunction towers, such as group completion, localization, and completion. The mapping hierarchy is implemented as a filtered 2-category, and time is given as a thin category (for example, $(\mathbb{N}, \leq)$ ). These categorical structures can, when necessary, be elevated to higher-dimensional forms such as n-, ∞-, or n-fold categories (e.g., Leinster, 2004; Lurie, 2009), or they are expected to be intuitively so; however, in this study, to avoid the complexity of higher coherence, the implementation is limited to 2-categorical structures.

3. **Definition of "Human" using IMU + HSG**
   This section presents the general framework for defining "human" using IMU + HSG. In particular, for the nervous system, a neural HSG based on neuron functions as a fundamental element is constructed. Here, the neuron function is built from the numerical system HSG as a composite of dendritic (fuzzy) and axonal (Boolean) functions. Based on this, a neural HSG is defined as a fuzzy logical function space capable of handling mappings and higher-order predicates. Furthermore, partial human functional universes such as glial, learning, endocrine, input–output, genetic, and material systems are established in parallel, managed through IMU, and the prospect is shown of describing behavior through inter-universal algorithmic compositions along temporal progression. The material system here serves as the base category shared across the entire universe, corresponding to mapping hierarchy 0, and the definition of human as a whole is given as a fibered-category construction. In addition, self-extending algorithms, facilitated by the functions of learning and genetic systems, as well as external–internal algorithms, enabled by the functions of input–output systems, are discussed, demonstrating that these lead to the logical construction necessary for defining AGI.

**Significance of this study**

By focusing on CES, IMU, and HSG, this study presents a comprehensive protocol of axioms that enables a complete separation between notation and meaning, connecting inter-universal system construction, human definition, and AGI design within a single framework. In other words, the structure of this study aligns with machine language in that it manages axiomatic systems as packages and treats notation as a language specification. Moreover, CES, notational axioms, and cumulative axioms are new and significant structures not only for applications but also for the foundations of mathematical logic itself. In particular, CES is considered directly applicable to formal logic beyond mathematics as well as to applied and philosophical sciences, and thus has a wide range of applicability. For example, constructing a CES-based system in natural language is trivially possible, and in that sense, this study is capable of bridging across all possible definitions.

## 1. Introduction of CES and the mathematical structure of IMU

The IMU introduced by Itoh (2025.08) indicates the necessity of inter-universal logical structures (e.g., Lurie, 2009; Mochizuki, 2012). When the concept of "inter-universal" is reinterpreted, for instance, as "between axiom systems," constructing logic starting from a specific mathematical system such as ZFC or HoTT would likely lead to opacity and redundancy in the theory. To overcome this axiomatic constructional problem, the present study introduces the *Cogito, ergo sum* axiom (CES) (Descartes, 1637). CES is defined as follows.

In the *Cogito, ergo sum* axiom (CES), "existence $E$" is defined such that, for any symbol $t$, a reflexive self-identity $S(t) = t$ holds by means of a reflective referential operation $S$. That is,

$$Cogito, Ergo\ Sum\ \text{Axiom} \coloneqq \big(E(t) \coloneqq (S(t) = t)\big) \tag{1}$$

Here, $S$ is a unary reflective referential operator representing reference, mention, or recursion within syntax, and this definition does not depend on any observer or external system. It gives existence as self-identity under referential possibility. The intended meaning here is that existence is not merely "being visible," but rather a syntactic stability in which "it can be stated," and "what is stated coincides with the original symbol." Furthermore, CES in this paper does not restrict the reflexive reference $S$ applied to the definition itself ($E(t) \coloneqq (S(t) = t)$) or to its constituting symbols. That is, it is posited as a minimal structure that "merely allows" reflexive reference, including self-application, self-inclusion, and iterative application (for example, $E\big(E(t)\big) \coloneqq (S\big(E(t)\big) = E(t))$). The responsibility of CES lies solely in this minimal nucleus of reflexive reference, and logical or mathematical consistency work, such as examining paradoxes or hypothetical minimal sets, is essentially placed outside the scope of CES itself. Constraints, consistencies, or typings necessary for such considerations are to be introduced as additional extensions to CES when appropriate. Therefore, CES in this study is the "axiom of axioms," replacing the internalized "I think" with the external and syntactic condition of "it can be said, it coincides, therefore it exists." Systems such as ZFC, HoTT, and natural language frameworks are positioned as attachable extension packs on top of this CES. The most convenient point of CES in logical construction is that, beyond conventional axiom systems, it allows freely constructing flexible axiom systems corresponding to any problem setting. For instance, although the following discussion employs mathematical logic, the CES-based logic can evidently be applied to natural language, computer languages, or even non-linguistic logic, thereby demonstrating its flexibility.

By decomposing the phrase "I think, therefore I am," each component can be mapped to the expression $E(t) \coloneqq (S(t) = t)$ as follows:

I: $t$,

think: $S()$,

therefore: $=$, and

am: $E() \coloneqq$.

Here, "therefore" is not interpreted as an inferential syntactic relation but is instead assigned the symbol "=" as an expression of the identity within reflexive reference. The reason why "am" corresponds to the syntactically more complex form $E() \coloneqq$ is that it reflects the ontological implication of CES—namely, that "am" is to be read as "to be defined as existing." Making the ontological implication of CES explicit within syntax reflects the intention to minimize subjective interpretation by external systems, the most representative example being human cognition itself, thereby making the resulting axiom system more non-personal and objective. It is possible to define the CES axiom simply as $S(t) = t$ and interpret "=" as "therefore … am" but this would easily be taken as imposing an ontological implication on the equality symbol or delegating that implication to external recognition, which is why this study does not adopt that form. However, if one intentionally embeds ontological implication into the equality symbol "=," or if CES is expressed using another symbol such as $\vdash$ or $\Rightarrow$ instead of "=," then a form like $S(t) =$ or $\vdash$ or $\Rightarrow t$ alone may also be adopted. In other words, some other minimal axiom could replace the CES axiom, and the CES system may be regarded as an example of an axiom system that could be called a "minimal axiom system." For instance, one could consider an opposite-phase construction that begins from "unreferability," as in Platonic idealism or Kantian noumenalism, and builds what is "referable" from there—an inverse approach to CES. Of course, there is no necessity to adhere to philosophical discourse itself; the choice of a minimal axiom will depend on the usefulness of the axiom system's construction. Developing the logic of such minimal axiom systems would, in turn, refine CES itself and contribute to the creation of more advanced axiomatic structures.

Let the minimal configuration of IMU, $\mathrm{IMU}_0$, consist solely of CES. That is,

$$\mathrm{IMU}_0 \coloneqq \{E(t) \coloneqq (S(t) = t)\} \tag{2}$$

and each axiom system, such as ZFC, is attached to it as an Axiom Package (AP):

$$\mathrm{IMU} \coloneqq \mathrm{IMU}_0 \cup \{\mathrm{AP}_1, \mathrm{AP}_2, \dots, \mathrm{AP}_\mathrm{n}\} \tag{3}$$

Including HSG discussed later, it is considered necessary in the CES framework to explicitly introduce "notational axioms" as part of the axioms. A notational axiom, strictly speaking, refers to all notations—characters, symbols, graphs, etc.—and their corresponding meanings used for representation. In other words, even the individual symbols and combinations in equation (1), as well as their semantic correspondences, are explicitly introduced as axioms. In conventional axiom systems, these are usually treated as implicit or external concepts. However, in the CES framework, notation-related axioms are also explicitly added, with CES itself remaining the minimal structure. This approach eliminates ambiguity in notation itself and, by doing so, makes it easier to incorporate the CES framework into computer or machine-proof languages. However, for simplicity, this paper does not define each individual symbol, self-evident notation, or axiom in sequence, but does so only for new axiomatic structures.

For the signature of an axiom system defined within IMU or for the contents defined by it, one may use, for example, institutional model theory (e.g., Goguen and Burstall, 1992). Performing this correspondence simultaneously is considered helpful for inter-universal theoretical construction. For instance, let ZFC and HoTT be denoted as $\mathcal{I}_{\mathrm{ZFC}}$ and $\mathcal{I}_{\mathrm{HoTT}}$, respectively, and in the institutional framework, these can be written as

$$\mathcal{I} = (\mathrm{Sign}, \mathrm{Sen}, \mathrm{Mod}, \vDash)$$

as an institution. Then,

$$\mathcal{I}_{\mathrm{ZFC}} \coloneqq (\mathrm{Sign}_\mathrm{Z}, \mathrm{Sen}_\mathrm{Z}, \mathrm{Mod}_\mathrm{Z}, \vDash_\mathrm{Z}), \mathcal{I}_{\mathrm{HoTT}} \coloneqq (\mathrm{Sign}_\mathrm{H}, \mathrm{Sen}_\mathrm{H}, \mathrm{Mod}_\mathrm{H}, \vDash_\mathrm{H}).$$

This allows axioms with model-theoretic signatures to be added to IMU. The correspondence between these can be implemented as a decorated institution (e.g., Diaconescu, 2008), and expressed as

$$F: \mathcal{I}_{\mathrm{ZFC}} \longrightarrow \mathcal{I}_{\mathrm{HoTT}} \text{ with } F = (\phi_\Sigma, \phi_{\mathrm{Sen}}, \phi_{\mathrm{Mod}}, \phi_\vDash),$$

imposing satisfaction-preservation conditions such as

$$M \vDash_Z \phi \Rightarrow \varphi_{\mathrm{Mod}}(M) \vDash_H \phi_{\mathrm{Sen}}(\phi)$$

when necessary. Nevertheless, in CES+IMU, only CES constitutes the minimal configuration, and all model-theoretic typings, including institutions, are additional axioms. Therefore, such frameworks should be introduced only when necessary in the logical construction.

## 2. Definition of HSG

### 2.0. Minimal configuration of HSG

In this section, we attempt to formalize the HSG introduced by Itoh (2025.08) by reintroducing it into the CES+HSG framework using categorical formalism. First, the minimal configuration of HSG ($\mathrm{HSG}_0$) is defined by its "set of axes," "index set for each axis," and "Cartesian product grid with tuple projection equalities (Cartesian notation)," which is introduced as a notational axiom within IMU:

$$\mathrm{HSG}_0 \coloneqq \left( \mathcal{A}, (I_a)_{a\in\mathcal{A}}, \prod_{a\in\mathcal{A}} I_a, \left( \pi_a : \prod_{a\in\mathcal{A}} I_b \to I_a \right)_{a\in\mathcal{A}} \right) \tag{4}$$

Here, $\mathcal{A} = \{a_i\}_i$ is the set of axes, each $I_a$ is an index set on axis $a$, $\prod_{a\in\mathcal{A}} I_a$ is the Cartesian product grid (state coordinate space), and $\pi_a$ is the Cartesian projection. Furthermore, this study adds a definability predicate as an auxiliary structure to $\mathrm{HSG}_0$:

$$\delta: \prod_{a\in\mathcal{A}} I_a \to \{\bot, \top\} \tag{5}$$

Thus, the extended version of $\mathrm{HSG}_0$ is defined as

$$\widehat{\mathrm{HSG}_0} \coloneqq (\mathrm{HSG}_0, \delta).$$

Its operational meaning is arbitrary depending on the object under consideration. A concrete example will be provided later in relation to definability over time. For instance, $\delta(x) = \top$ represents that $x$ is "definable," while $\delta(x) = \bot$ の represents that $x$ is "undefinable," thereby offering a meta-level parameterization.

The formulation above differs from that in Itoh (2025.08) in that it separates the notational and content components, taking the former as the minimal structure and leaving the latter free. This approach enhances the freedom and notational flexibility of HSG as an implementation concept, allowing it to be introduced as a maximally general formal framework. That is, the setting of axes in HSG does not necessarily need to be state depth, mapping hierarchy, or time, and the number of axes is not limited to three. Such an implementation of HSG enables constructions with greater freedom, as demonstrated in Chapter 3, for example, in the implementation of the human definition as a fibered category combining parallel subcategories with a shared base category.

### 2.1. Definition of HSG using category theory

Next, the numerically structured HSG of Itoh (2025.08) is reimplemented in a mathematically formalized form within the categorical framework. For the set of axes of the numerical system HSG $\mathcal{A} = \{a_i\}_i$, each axis is regarded as a 2-category $C_i$, and the whole system is given as their product 2-category:

$$\mathcal{H} \coloneqq \prod_{i\in\mathcal{A}} \tau_{\leq 2}(C_i) \tag{6}$$

Then, following $\mathrm{HSG}_0$, the numerical system HSG is constructed in Cartesian coordinate notation.

#### 2.1.1. State depth: the category of cumulative axioms

The state-depth axis is constructed as a "cumulative-axiom" 2-category in which axioms are successively layered, connecting each depth through adjunctions of free construction/forgetful or classification/representation:

$$\mathcal{C}_{\text{depth}} \coloneqq C_0 \overset{\dashv_J}{\to} C_1 \xrightarrow{F_1 \dashv G_1} \cdots \xrightarrow{F_{n-1} \dashv G_{n-1}} C_n \tag{7}$$

Here, each $F_i \dashv G_i (i \geq 1)$ denotes an adjunction such as free construction/forgetful or reflection, giving the addition or classification of axioms toward higher structures and the forgetting or representation toward lower ones. Between $C_0$ and $C_1$, however, no direct internal adjunction is placed; instead, they are connected through a $J$-relative pseudo-adjunction $\dashv_J$ mediated by an external criterion $J$. A concrete example of the state-depth sequence in the numerical system HSG is as follows:

$$\frac{\text{Undef}}{\text{Define}} \to \frac{\text{Empty}}{\text{NonEmpty}} \to \text{Set} \to \text{OrdSet} \to \text{Ring} \to \text{Field} \to \text{Complete Field} \tag{8}$$

The categorical structures and adjunctions for each depth are defined below. In particular,

$$C_0\left(\frac{\text{Undef}}{\text{Define}}\right) \overset{\dashv_J}{\to} C_1\left(\frac{\text{Empty}}{\text{NonEmpty}}\right) \xrightarrow{F_1 \dashv G_1} C_2\ (\text{Set})$$

contains many concepts originally introduced in Itoh (2025.08), and is described in detail here. From equation (5),

$$\delta: \prod_{a \in \mathcal{A}} I_a \to \{\bot, \top\}$$

we define the definable objects universe $X_\delta$, including $\mathcal{C}_{\text{depth}}$ as

$$X_\delta \coloneqq \text{dom}(\delta) = \prod_{a \in \mathcal{A}} I_a.$$

We then define a thin category $\mathcal{T}_\delta$ as a preordered version of this:

$$\text{Ob}(\mathcal{T}_\delta) = X_\delta, \text{Hom}_{\mathcal{T}_\delta}(x, y) = \begin{cases} \{*\}\ \delta(x) \leq \delta(y)\ (\bot < \top) \\ \quad \emptyset \text{ otherwise} \end{cases}$$

If necessary, representatives $\bot_\delta$ or $\top_\delta$ are freely adjoined when $\delta^{-1}(\bot)$ or $\delta^{-1}(\top)$ is empty. Then, $\tilde{\delta}: \mathcal{T}_\delta \to 2 = \{\bot \to \top\}$ becomes a functor, since arrow preservation follows immediately from the monotonicity of $\leq$.

#### 2.1.1.1. $C_0\left(\frac{\text{Undef}}{\text{Define}}\right)$

The category $C_0\left(\frac{\text{Undef}}{\text{Define}}\right)$ is defined as a two-point preorder category:

$$C_0 \coloneqq \{\text{Undef} \to \text{Define}\}.$$

The classification functor between $\mathcal{T}_\delta$ and $C_0$ is defined as

$$\Sigma^\delta_{\text{def}}: \mathcal{T}_\delta \to C_0, \Sigma^\delta_{\text{def}}(x) = \begin{cases} \text{Undef} & \delta(x) = \bot \\ \text{Define} & \delta(x) = \top \end{cases},$$

and the representation functor as

$$\iota_0^\delta : C_0 \to \mathcal{T}_\delta, \iota_0^\delta(\mathrm{Undef}) = \bot_\delta, \iota_0^\delta(\mathrm{Define}) = \top_\delta.$$

These form an adjunction:

$$\Sigma_{\mathrm{def}}^\delta \dashv \iota_0^\delta,$$

with the natural isomorphism

$$\mathrm{Hom}_{C_0}\left(\Sigma_{\mathrm{def}}^\delta x, a\right) \cong \mathrm{Hom}_{\mathcal{T}_\delta}\left(x, \iota_0^\delta a\right).$$

This holds because the left-hand side is nonempty exactly when $\delta(x) \leq \hat{a}$, and the right-hand side, $x \to \iota_0^\delta(a)$, is equivalent to the same condition. Its naturality is immediate since $\mathcal{T}_\delta$ is thin. The unit and counit are given by

$$\eta_x : x \to \iota_0^\delta \Sigma_{\mathrm{def}}^\delta x, \varepsilon_a : \Sigma_{\mathrm{def}}^\delta \iota_0^\delta a \to a,$$

and the corresponding monad

$$T_0 \coloneqq \iota_0^\delta \Sigma_{\mathrm{def}}^\delta$$

is idempotent (reflective). When smallness is considered, it suffices to take $X_\delta$ within a Grothendieck universe. Note that this adjunction $\Sigma_{\mathrm{def}}^\delta \dashv \iota_0^\delta$ is purely axiomatic—it does not belong to the categorical structures of Eq. 7 or 8 but rather connects the definability tag with the universe itself. Based on the CES+IMU principle of explicitness and elimination of ambiguity, setting such a definable universe is regarded as an unavoidable operation.

#### 2.1.1.2. $C_1\left(\frac{\mathrm{Empty}}{\mathrm{NonEmpty}}\right) \xrightarrow{F_1 \dashv G_1} C_2\ (\mathbf{Set})$

The category $C_1(\frac{\mathrm{Empty}}{\mathrm{NonEmpty}})$ is a two-point preorder category:

$$C_1 \coloneqq \{\mathrm{Empty} \to \mathrm{NonEmpty}\},$$

and $C_2(\mathrm{Set})$ is the category of sets without internal ordering:

$$C_2 \coloneqq \mathrm{Set}.$$

Define the representation functor between $C_1$ and $C_2$:

$$\iota_1 : C_1 \to C_2, \iota_1(\mathrm{Empty}) = \emptyset, \iota_1(\mathrm{NonEmpty}) = 1,$$

and the classification functor:

$$\Sigma_{\mathrm{emp}} : C_2 \to C_1, \Sigma_{\mathrm{emp}}(A) = \begin{cases} \mathrm{Empty}\ A = \emptyset \\ \mathrm{NonEmpty}\ A \neq \emptyset \end{cases}.$$

Then the adjunction is constructed as

$$F_1 \dashv G_1 \coloneqq \Sigma_{\mathrm{emp}} \dashv \iota_1,$$

with

$$\mathrm{Hom}_{C_1}(\Sigma_{\mathrm{emp}}A, b) \cong \mathrm{Hom}_{C_2}(A, \iota_1 b)$$

naturally holding. Consequently, the induced monad

$$T_1 \coloneqq \iota_1\Sigma_{\mathrm{emp}}: \mathrm{Set} \to \mathrm{Set}, \mathrm{T}_1(A) = \begin{cases} \emptyset\ (A = \emptyset) \\ 1\ (A \neq \emptyset) \end{cases}$$

is idempotent (reflective). This construction provides a minimal Boolean-style definition of existence—collapsing nonempty sets to a single point and fixing empty ones to $\emptyset$.

### 2.1.1.3. $\boldsymbol{C_0}\left(\frac{\mathbf{Undef}}{\mathbf{Define}}\right) \overset{\dashv_J}{\to} \boldsymbol{C_1}\left(\frac{\mathbf{Empty}}{\mathbf{NonEmpty}}\right)$: Quasi-adjunction via external criterion $\boldsymbol{J}$

The connection between $C_0$ and $C_1$ is, as shown in Itoh (2025.08), not a structure closed within a single universe but a quasi-adjunction mediated by an inter-universal external criterion $J$. This quasi-adjunction establishes the semantic syntax of "what is taken as definable," forming the core for defining meta-concepts such as contradiction/consistency, as well as observed/unobserved, and verified/unverified. Let the external criterion be defined as

$$J: \mathcal{T}_\delta \to \mathrm{Set} \tag{9}$$

assigning to each token $x \in \mathcal{T}_\delta$ a carrier set labeling its definability. Here, Set is not the internal Set category $C_2$ but a meta-level Set on the IMU side, used to assign definability externally, and $J$ is a classification functor into it.

Operationally, this can be realized as a guarded semantics syntax:

$$J(x) \coloneqq \begin{cases} [\![x]\!]\ \mathrm{guard}(x)\ \mathrm{holds} \\ \emptyset\ \mathrm{guard}(x)\ \mathrm{fails} \end{cases},$$

where $\mathrm{guard}(x)$ is built from concepts such as "no future reference," "constraint," "type consistency," or "authorization."

Define the representative functor Tr in the direction $C_0 \to C_1$:

$$\mathrm{Tr}: C_0 \to C_1, \mathrm{Tr}(\mathrm{Undef}) = \mathrm{Empty}, \mathrm{Tr}(\mathrm{Define}) = \mathrm{NonEmpty}.$$

To ensure coherence between $J$ and Tr, define a comparison natural transformation:

$$\kappa^\delta: \Sigma_{\mathrm{emp}} \circ J \Rightarrow \mathrm{Tr} \circ \Sigma_{\mathrm{def}}^\delta,$$

where each component $\kappa_x^\delta$ is a morphism (either identity or $\mathrm{Empty} \to \mathrm{NonEmpty}$) in $C_1$. Thus, the consistency and soundness conditions for the external criterion J can be defined as follows:

$$(\mathrm{J0})\ \delta(x) = \bot \Rightarrow J(x) = \emptyset$$

$$(\mathrm{J1})\ \delta(x) = \top \Rightarrow J(x) \neq \emptyset$$

Under condition (J0), $\kappa^\delta$ always exists, and under both (J0) and (J1):

$$\mathrm{Hom}_{C_1}(\Sigma_{\mathrm{emp}}J(x), b) \cong \mathrm{Hom}_{C_0}(\Sigma_{\mathrm{def}}^\delta x, \mathrm{Tr}b)$$

naturally holds for all $x, b$. Hence,

$$\Sigma_{\mathrm{emp}} \circ J \dashv_J \mathrm{Tr} \circ \Sigma_{\mathrm{def}}^{\delta}$$

constitutes a $\mathrm{J}$-relative quasi-adjunction. When only $(\mathrm{J0})$ holds, the correspondence is not an isomorphism but an implication:

$$\mathrm{Hom}_{C_1}\left(\Sigma_{\mathrm{emp}} J(x), b\right) \Rightarrow \mathrm{Hom}_{C_0}\left(\Sigma_{\mathrm{def}}^{\delta} x, \mathrm{Tr} b\right).$$

As mentioned, the interpretation of the terms $\mathrm{Empty/NonEmpty}$ in $\Sigma_{\mathrm{emp}} J(x)$ depends on the design of $J$. Examples include:

- **Observation semantics:** Empty = unobserved, NonEmpty = observed.
- **Reachability:** Empty = unreachable, NonEmpty = reachable.
- **Authorization:** Empty = unauthorized, NonEmpty = authorized.

The necessity of introducing a minimal axiomatic system including CES becomes evident in this $J$-quasi-adjunction. The $J$-quasi-adjunction is inherently inter-universal, defined across multiple heterogeneous axiom systems (e.g., ZFC, HoTT, ETCS, or natural language-based theories). In such inter-universal settings, fixing one theory (such as ZFC) as a base and embedding others within it generally leads to non-canonical and complex constructions, risking the loss of essential meaning and structural consistency.

To avoid this, it is effective to establish a lightweight, neutral foundation independent of any particular system—a *minimal axiom system*. Each axiom system is then related via translations into this minimal framework, allowing categorical operations such as comparison, composition, and combination while preserving inter-theoretic coherence. Thus, implementing categorical constructions like J-quasi-adjunctions in a multi-theoretic logical space factually requires such a neutral reference point, and CES and similar minimal axioms precisely fulfill this role. Therefore, within IMU's inter-universal logical construction, the minimal axiom system is not merely a theoretical option but a methodologically unavoidable starting point.

#### 2.1.1.4. Set → OrdSet → Ring → Field → Complete Field

From depth level 2 onward, structures are successively layered—from sets to ordered sets, rings, fields, and finally complete fields. Each layer is connected by an adjoint pair consisting of a left adjoint that freely adds structure and a right adjoint that forgets it, ultimately leading to the construction of a complete field.

(1) $\mathrm{Set} \to \mathrm{OrdSet}$ (e.g., Awodey, 2006)

The transition from sets to ordered sets is defined by a free construction that assigns a discrete order and a right adjoint that forgets the order:

$$\Delta: \mathrm{Set} \to \mathrm{OrdSet}, U_{\mathrm{pos}}: \mathrm{OrdSet} \to \mathrm{Set}, F_2 \dashv G_2 \coloneqq \Delta \dashv U_{\mathrm{pos}}$$

Here, $\Delta(X) = (X, =)$ is the free construction assigning a discrete order, and $U_{\mathrm{pos}}$ extracts the underlying set by forgetting the order structure.

(2) $\mathrm{OrdSet} \to \mathrm{Ring}$ (e.g., Leinster, 2016)

For the transition from ordered sets to rings, a free commutative ring is assigned to the underlying set, and the right adjoint returns to the discrete order:

$$F_{\mathrm{ring}}\colon \mathrm{OrdSet} \to \mathrm{CRing}, G_{\mathrm{ring}}\colon \mathrm{CRing} \to \mathrm{OrdSet}, F_3 \dashv G_3 \coloneqq F_{\mathrm{ring}} \dashv G_{\mathrm{ring}}$$

Using the forgetful functor $U_{\mathrm{set}}\colon \mathrm{CRing} \to \mathrm{Set}$, we set $F_{\mathrm{ring}}(P) = \mathrm{FreeRing}(U_{\mathrm{pos}}(P)), G_{\mathrm{ring}}(R) = \Delta(U_{\mathrm{set}}(R))$, so that the free assignment and the forgetting of ring structure hold naturally.

(3) $\mathrm{Ring} \to \mathrm{Field}$ (via integral domains) (e.g., Leinster, 2004; Riehl, 2016)

The transition from commutative rings to fields is given as a reflection through the subcategory of integral domains. Let $\mathrm{Dom} \subset \mathrm{CRing}$ be the subcategory of integral domains. Then:

$$\mathrm{Frac}\colon \mathrm{Dom} \to \mathrm{Field}, U_{\mathrm{dom}}\colon \mathrm{Field} \to \mathrm{Dom}, F_4 \dashv G_4 \coloneqq \mathrm{Frac} \dashv U_{\mathrm{dom}}$$

Here, $\mathrm{Frac}(R)$ is the field of fractions of an integral domain $R$, and any ring homomorphism $R \to U_{\mathrm{dom}}(K)$ into a field $K$ uniquely extends to $\mathrm{Frac}(R) \to K$.

(4) $\mathrm{Field} \to \mathrm{Complete\ Field}$ (by completion) (e.g., Kelly, 2005)

Completion of fields is defined as a reflection, fixing a topology (or valuation):

$$\widehat{(-)}\colon \mathrm{Field}_v \to \mathrm{CField}_v, U_{\mathrm{comp}}\colon \mathrm{CField}_v \hookrightarrow \mathrm{Field}_v, F_5 \dashv G_5 \coloneqq \widehat{(-)} \dashv U_{\mathrm{comp}}$$

Here, $\widehat{K}$ denotes the valuation-complete field of $K$, equipped with the natural embedding $\eta_K\colon K \to U_{\mathrm{comp}}(\widehat{K})$. This reflection is idempotent—once completed, the structure stabilizes ($\widehat{\widehat{K}} \simeq \widehat{K}$).

Summarizing the above, the higher layers of state depth in the numerical system HSG form the following tower of adjunctions:

$$\mathrm{Set} \xrightarrow{\Delta \dashv U_{\mathrm{pos}}} \mathrm{OrdSet} \xrightarrow{F_{\mathrm{ring}} \dashv G_{\mathrm{ring}}} \mathrm{Ring} \xrightarrow{\mathrm{Frac} \dashv U_{\mathrm{dom}}} \mathrm{Field} \xrightarrow{\widehat{(-)} \dashv U_{\mathrm{comp}}} \mathrm{Complete\ Field} \qquad (10)$$

This tower connects the hierarchical structures of order, algebra, field, and completeness through free/forgetful adjunctions, beginning from the set-theoretic foundation. It defines the upper structural layers of state depth in the numerical system HSG. Thus, the state-depth axis of HSG can be understood as a consistent chain of adjunctions running from definability up to completeness, providing a categorical framework that unifies all hierarchical levels of definition, existence, structure, and completion.

### 2.1.2. Mapping-hierarchy axis: Free/forgetful filtration as dimensional layering

The mapping-hierarchy axis is defined as a filtered 2-category, in which higher-order morphisms are freely added stepwise within a single object universe, and forgotten in the reverse direction:

$$C_{\mathrm{map}} \coloneqq \{C^k\}_{k \in \mathbb{N}}, \qquad i_k\colon C^{(k)} \hookrightarrow C^{(k+1)} \text{ (id on objects, locally full)} \qquad (11)$$

At each stage, a free/forgetful adjunction is made explicit:

$$F_k^{\mathrm{map}}\colon C^{(k)} \to C^{(k+1)}, U_k^{\mathrm{map}}\colon C^{(k+1)} \to C^{(k)},$$

$$F_k^{\mathrm{map}} \dashv U_k^{\mathrm{map}}$$

Here, $i_k$ is the canonical inclusion (identity on objects, locally full) into the image of $F_k^{\mathrm{map}}$.

Intuitively, $C^{(0)}$ (mapping level 0) has only objects (a discrete 2-category), $C^{(1)}$ freely adds 1-morphisms, $C^{(2)}$ freely adds 2-morphisms, and higher levels add filtered higher morphisms. Although the mapping hierarchy could be extended to n-categories, we restrict it to a filtered 2-category to avoid higher-coherence complexity. When necessary, the filtered colimit is fixed as a colimit (depending on context) in 2-Cat or Cat:

$$C^{(\infty)} := \mathrm{colim}_k C^{(k)}$$

Thus, the mapping-hierarchy axis builds dimensional filtration within a single category, while the state-depth axis connects distinct categories $C_i$ via free/forgetful adjunctions $F_i \dashv G_i$ — an axiomatic filtration. The former describes higher-dimensional morphism structures within one categorical universe; the latter provides depth across different categorical layers.

**2.1.3. Time axis: Thin category and prohibition of future reference**

Time is represented as a thin category of natural numbers with their standard order, and the projection in $\mathrm{HSG}_0$ gives its coordinate:

$$\mathcal{T} := (\mathbb{N}, \leq), a_\tau \in A, I_{a_\tau} = \mathbb{N}, \tau := \pi_{a_\tau}: X_\delta \to \mathbb{N} \quad (12)$$

Define a dependency relation $\mathrm{dep} \subseteq X_\delta \times X_\delta$, where $(x, y) \in \mathrm{dep}$ means "$x$ depends on $y$."

We formalize the prohibition of future reference ("no reference to future states") in two equivalent ways by Eq. 5 and §2.1.1.3:

(A) δ-based formulation

$$\exists y\big((x, y) \in \mathrm{dep} \wedge \tau(y) > \tau(x)\big) \Rightarrow \delta(x) = \bot \quad (13)$$

This ensures that "an element depending on a future one is undefined," structurally guaranteed within HSG.

(B) External-criterion ($J$-based) formulation

For each time $t \in \mathbb{N}$, define an external criterion $J_t$:

$$J_t: \mathcal{T}_\delta \to \mathrm{Set}, J_t(x) = \begin{cases} [\![x]\!]_t & x \text{ can be evaluated using only data with } \tau(y) \leq t \\ \emptyset & \text{otherwise} \end{cases}$$

As before, Set here is the external Set in IMU, not $C_2$'s internal one.

The soundness $(J0)$ and sufficiency $(J1)$ conditions,

$$(J0)\ \delta(x) = \bot \Rightarrow J_t(x) = \emptyset$$

and

$$(J1)\ \delta(x) = \top \Rightarrow J_t(x) \neq \emptyset$$

may be influenced by other properties of J, so they are not fully formalized here.

**2.1.4. Axiomatization of "Definition = State"**

Following Itoh (2025.08), we formalize the principle of "*definition* = *state*" in HSG as a statement of syntactic uniqueness. Define the subuniverse of defined elements:

$$\mathrm{Def}_\delta \coloneqq \{x \in X_\delta | \delta(x) = \top\} \subseteq X_\delta$$

For each projection $\pi_a$, restrict it to

$$\pi_a|_{\mathrm{Def}_\delta} : \mathrm{Def}_\delta \to I_a.$$

We then introduce the unique determination by coordinates as an axiom:

$$\forall x, y \in \mathrm{Def}_\delta, \left(\forall a \in A, \pi_a(x) = \pi_a(y)\right) \Rightarrow x = y.$$

Equivalently, the restricted product projection

$$\langle \pi_a|_{\mathrm{Def}_\delta} \rangle : \mathrm{Def}_\delta \to \Pi_{a \in A} I_a \text{ is injective.}$$

Thus, being "defined" ($\delta(x) = \top$) means that an element $x$ has well-defined coordinate values $\pi_a(x) \in I_a$ on all axes $a \in A$, and this tuple uniquely specifies $x$. Inside the axiomatic structure of HSG, "state" refers to members of $\mathrm{Def}_\delta$; hence, every definable element $x$ corresponds precisely to a unique state. Consequently, "being defined" and "being a state" coincide perfectly within HSG, constituting the formal syntactic expression of the principle "definition = state." If none of the axes are defined (i.e. $\delta(x) = \perp$), then $x \notin \mathrm{Def}_\delta$ and is considered undefined/unobserved as a state. This reflects that HSG need not be complete with respect to observation—a conceptual allowance for partiality within definability.

### 2.2. Numerical System HSG Table

Based on the definitions so far, Table 2 in Itoh (2025.08) is reintroduced here as Table 1. For details on the content and specific operations defined on each grid, refer to Itoh (2025.08).

| | | | | | | | |
|---|---|---|---|---|---|---|---|
| $C^{(3)}$ | | Higher-order logic | Set properties | Order properties | Algebraic properties | Field properties | Completeness properties |
| $C^{(2)}$ | | Logical predicates | Set predicates | Order predicates | Ring predicates | Field predicates | Completeness predicates |
| $C^{(1)}$ | Definition judgment maps | Truth-value functions | Set morphisms | Order morphisms | Ring homomorphisms | Field homomorphisms | Complete homomorphisms |
| $C^{(0)}$ | Definability | Empty/NonEmpty | Set | OrdSet | Ring | Field | Complete Field |
| **Mapping hierarchy / State depth** | $C_0$ | $\xrightarrow{\dashv_J}$ $C_1$ | $\xrightarrow{F_1 \dashv G_1}$ $C_2$ | $\xrightarrow{F_2 \dashv G_2}$ $C_3$ | $\xrightarrow{F_3 \dashv G_3}$ $C_4$ | $\xrightarrow{F_4 \dashv G_4}$ $C_5$ | $\xrightarrow{F_5 \dashv G_5}$ $C_6$ |

**Table 1.** Reintroduction of the numerical system HSG from Itoh (2025.08). Each grid corresponds to the relevant adjunctions connecting categorical levels.

## 2.3 Kan Extensions in HSG

To verify the naturality of HSG and the states defined within it, one should confirm that the entire construction satisfies Kan extension properties. In essence, this means checking—pointwise—whether HSG's constructions extend uniquely and naturally under changes of axes (depth, mapping hierarchy, time) or bases (forgetful/free adjunctions, J-relative quasi-adjunctions). Left Kan extensions correspond to minimal generation (free addition), while right Kan extensions correspond to restriction without information loss (safe forgetting). If these can be attached to each layer, naturality can be guaranteed globally—meaning that "the outcome of the definition is automatically determined."

Intuitively, the correspondence can be understood as follows. Along the state-depth axis, each ascent (Set→OrdSet→Ring …) is a left Kan extension (free construction), while the reverse forgetting is a right Kan extension. Along the mapping-hierarchy axis, showing that each inclusion $C^{(k)} \hookrightarrow C^{(k+1)}$ induces pointwise left/right Kan extensions for "adding/forgetting higher morphisms" suffices. Along the time axis, the prohibition of future reference can be interpreted as a restriction to the past—a right Kan extension expressing universal restriction—while monotonicity over time (more information ⇒ greater evaluability) follows the same right-Kan pattern. The J-relative adjunction between $C_0$ and $C_1$ can likewise be organized as a relative (lax/oplax) Kan extension, ensuring that internal constructions remain invariant even when the external criterion J is varied—comparison natural transformations become unique.

Practically, the verification is reduced to three conditions. (i) Each layer's free construction, forgetful functor, fraction field formation, and completion can be computed as pointwise left/right Kan extensions. (ii) On the product of axes, a Beck-Chevalley-type base-change condition (e.g., Mac Lane and Moerdijk 1992) holds so that glued results coincide. (iii) Smallness and the existence of (co)limits—including for $\mathrm{HSG}_0$ within any chosen universe—are ensured. If these hold, the transport of states and predicates in HSG becomes canonical and base-independent; reparametrization or changes of reference frame act via natural isomorphism. The formal proof should proceed, after standardizing CES (fixing axiomatic notation and vocabulary), by reconstructing each stage from the units and counits of the adjunctions as pointwise Kan extensions. In conclusion, by positioning every construction of HSG (including relative ones) as a Kan extension, the equivalence of "definition = state," the consistency between the depth-tower and the mapping hierarchy, and the temporal stability under constraints are all unified and guaranteed through categorical universality.

## 2.4. Remarks from a Learning-Oriented View of Category Theory

The categorical construction of HSG incidentally yields a structure that is not only theoretically rigorous but also pedagogically clear—potentially easier to grasp than conventional formulations. The intuitive factors contributing to this clarity can be summarized as follows:

(1) A visible map from the start.

Concepts have coordinates on the three axes—state depth, mapping hierarchy, and time—so every definition, proposition, or construction has a definite location; one never gets lost.

(2) Arrows have a consistent meaning.

Adjunctions are always read as adding (left) / forgetting (right).

Both the chain Set→OrdSet→Ring→… and the J-relative adjunction $C_0 \to C_1$ share this interpretation.

Even when new categories are introduced, the intuition “left = generate, right = forget” never breaks.

(3) Natural use of Kan extensions.

The intuition “extend minimally (left) / restrict safely (right)” absorbs every operation—free construction, forgetting, fraction formation, completion, and J-comparison—into a single unifying principle.

(4) Points of failure become visible.

Layered organization reveals exactly where Kan extensions fail—i.e., where naturality breaks. When a Kan extension exists, higher-order constructions automatically stabilize.

(5) Commutativity acts as a teacher.

Checking diagrammatic commutativity over the depth × hierarchy grid immediately tests the soundness of design: commutative = valid; non-commutative = assumption mismatch. This simplifies the learner’s diagnostic process.

(6) A consistent tactile experience.

Adding new objects never changes the procedure: add on the left, forget on the right, relativize if necessary, align via comparison, and finally collapse via Kan extension. The process is repeatable; notation grows, but the intuitive “feel” remains stable.

(7) “Definition = State” becomes intuitive.

Because the injectivity of the defined subuniverse is central, “to define = to select a point-state” is grasped directly. Even when $J$ is later imposed, this core intuition remains intact.

(8) Scalability to new theories.

Adding measure, topos, probability, geometry, or computational effects requires only deciding where on the axes the new objects live and which adjunction/relativization connects them. No redesign is needed—just a coordinate extension. This aligns with the Mac Lane-style assertion that “*every concept is a Kan extension* (Mac Lane, 1971).”

In short, HSG serves as a ladder whose rungs make clear which components are essential, which are optional, where freedom lies, and where constraints bind—both visually and procedurally. Because this ladder itself is built from categorical universality (adjunctions and Kan extensions), it preserves—and even strengthens—the conceptual essence of category theory while providing a structured aid for understanding and design.

## 3. Definition of "Human" using IMU + HSG

In this section, we present an overall framework for defining humans using the CES + IMU + HSG framework developed so far. The procedure is organized as follows: 3.1. construction of a neural HSG; 3.2. definition of human via multi-universe parallelization by fibered categorification; 3.3. discussion.

### 3.1. Construction of a neural HSG

We implement the nervous system—formed by neurons and their interconnections—on the time-indexed numerical-system HSG constructed in §2. The tower of adjunctions is set up in the order valued complete field → 0D neuron function body → 1D → 2D → 3D, with each stage connected, as before, by left: free construction/right: forgetting. In this paper, the valued complete field is fixed to $\mathbb{R}$, and we then fix the three-dimensional real vector space with the standard inner product, $E \coloneqq \mathbb{R}^3$. Henceforth, spatial coordinates always use $E$, and time uses $\mathbb{T} = (\mathbb{N}, \leq)$. Here, a neuron function is a function on three-dimensional Euclidean space obtained as a composite of a dendrite-like fuzzy function and an axon-like Boolean function. The neuron-function body can freely add higher dimensions as a complex-type construction with a structural composition in 0–3D.

#### 3.1.1 Neuron function

The neuron function defined in this study is a composite of a dendrite-like fuzzy function in the three-dimensional Euclidean space $E$ and an axon-like Boolean function; it is placed in $C^{(1)}$, i.e., as a 1-morphism. Its domain and codomain both lie in $C^{(0)}$, with carriers given by physical parameters such as chemical concentrations, membrane potentials, and receptor states. An input–output time lag is set and implemented in the time-indexed HSG. In $C^{(2)}$ and higher, we place abstract predicates and properties that take as their domain, for example, the neuron function itself or its firing patterns.

Let $P_{\text{in}}(r,t) \in C^{(0)}$ and $P_{\text{out}}(r,t) \in C^{(0)}$ denote the input and output parameter carriers at each point $(r,t) \in E \times \mathbb{T}$. Then, as a 1-morphism in the mapping hierarchy $C^{(1)}$, the neuron function is defined by

$$N_{r,t}: P_{\text{in}}(r,t) \to P_{\text{out}}(r, t+\Delta) \tag{14}$$

where $\Delta \in \mathbb{N}$ is a fixed time delay. By the prohibition of future reference (§2.1.3(A)), $N_{r,t}$ may refer only to information up to time $t$. While the model can freely choose the internal construction, a typical intuitive decomposition takes $\phi$ as a dendritic fuzzy-receptive function and $b$ as an axonal Boolean/threshold function, yielding the composite

$$P_{\text{in}}(r,t) \xrightarrow{\phi} [0,1]^m \xrightarrow{b} P_{\text{out}}(r, t+\Delta). \tag{15}$$

### 3.1.2 0D neuron-function body

By discretely placing these neuron functions on EEE, we construct a depth of complex-type structure with neuron functions as basic units. The 0D neuron-function body is a collection of placements of neuron functions, forming sequences of inputs and outputs as time progresses.

Assign to each finite countable point set $P \subset E$ and each point $p \in P$ the following objects of $C^{(0)}$ and $C^{(1)}$:

$C^{(0)}$: local carrier state $S_p$ (e.g., species concentrations, membrane potential),

$$C^{(1)}: N_p: P_{\text{in}}(p,t) \to P_{\text{out}}(p, t+\Delta).$$

We then define the category $\text{NF}_0$ of 0D neuron-function bodies by

$$\text{NF}_0 = \{(P, \{N_p, S_p\}_{p \in P}) | P \subset E\} \quad (16)$$

Let $\text{FinSet}_E$ denote the subcategory of finite countable discrete subsets in $E$. For $\text{FinSet}_E$ and $\text{NF}_0$, define the forgetful functor $U_N: \text{NF}_0 \to \text{FinSet}_E$ that returns the point set, and the free construction $F_N: \text{FinSet}_E \to \text{NF}_0$ that equips it with default neuron functions and carriers; then set the adjunction

$$F_N \dashv U_N.$$

This 0D depth $\text{NF}_0$ can be understood as the "minimal configuration of a nervous system that places neuron functions ($C^{(1)}$) on physical parameter carriers ($C^{(0)}$) located at point sets in Euclidean space $\mathbb{R}^3$."

### 3.1.3. Free addition from 0D to 3D (line, layer, and volume connections)

Whereas the 0D neuron-function body is a set of neuron functions, the 1–3D depths define depictions in which neurons form geometric structures of the corresponding dimensions. That is, the nervous system as a whole forms a 3D neuronal structure, while systems processing visual information function via 2D structures.

Let $\text{NF}_1, \text{NF}_2, \text{NF}_3$ denote the categories of 1–3D neuron-function bodies. Here, by using the skeleton $\text{Sk}_n$ as a truncation to dimension ≤n (forgetting information above n) and the coskeleton $\text{CoSk}_n$ as a uniquely determined extension from n-dimensional data to ≥n + 1 dimensions (e.g., Goerss–Jardine, 1999; Riehl, 2016), we construct the adjunction tower from $\text{NF}_0$ to $\text{NF}_3$:

$$\text{NF}_0 \xrightarrow{\text{Sk}_1 \dashv \text{CoSk}_1} \text{NF}_1 \xrightarrow{\text{Sk}_2 \dashv \text{CoSk}_2} \text{NF}_2 \xrightarrow{\text{Sk}_3 \dashv \text{CoSk}_3} \text{NF}_3 \quad (17)$$

When necessary, we use geometric realization $|-|: \text{sSet} \to \text{Top}$ and embed $|\text{NF}_n| \hookrightarrow E$ into space. The key point is that the skeleton/coskeleton act on the shape. Namely, for a fixed shape-generating functor

$$G: \text{FinSet}_E \to \text{sSet}, \quad P \longmapsto K_P \coloneqq G\ (P)$$

we apply $\text{Sk}_n \dashv \text{CoSk}_n$ to the resulting combinatorial shape $K_P$. Meanwhile, the neuron function $N_p$ and the carrier $S_p$ are treated as decorations placed on the shape and transported along these "shape adjunctions."

The structures $\text{NF}_0$ through $\text{NF}_3$ are then interpreted as follows:

$NF_0$: a collection of individual neurons (a state in which the nervous system is not yet identified as a network structure),

$NF_1$: a line network with synaptic connections added (e.g., the vector layer of a simple neural network),

$NF_2$: a surface network with layers/sheets added (e.g., visual cortex layers),

$NF_3$: a volume network with volumetric structure added (e.g., brain neural structures).

### 3.1.5. Neural-system HSG Table

Through the above construction, the neural system HSG can be represented as shown in Table 2. For simplicity, the time axis is not explicitly shown, though it should be understood that the system is time-indexed. The specific contents above $C^{(2)}$ will be discussed in §3.2 and beyond; however, since intra-human algorithms (composite morphisms) are described via inter-universal connections in the fibered construction, the predicates and properties that can be defined solely within the neural-system HSG are likely limited (§3.3.2).

| | | | | |
|---|---|---|---|---|
| $C^{(3)}$ | Point-neuron property | Line-neuron property | Surface-neuron property | Volume-neuron property |
| $C^{(2)}$ | point-neuron predicates | line-neuron predicates | surface-neuron predicates | volume-neuron predicates |
| $C^{(1)}$ | neuron function point | line-connected neurons | surface-connected neurons | volume-connected neurons |
| $C^{(0)}$ | local potential / neurotransmitter | vector potential / material distribution | matrix potential / material distribution | 3D tensor potential / material distribution |
| Mapping hierarchy / state depth | $\mathrm{NF}_0$ | $\mathrm{NF}_1$ | $\mathrm{NF}_2$ | $\mathrm{NF}_3$ |

**Table 2.** Neural-system HSG. $\mathrm{FinSet}_E$ and below are omitted.

### 3.2 Definition of "Human" via Multi-Universe Parallelization by Fibered Categorification (Fig. 2)

In this section, we extend the neural-system HSG to describe the definition of "human" through fibered categorification that combines multi-system parallelism with the shared material base $C^{(0)}$. The human definition aims to describe all human functions through a fibered human category $\mathcal{N}$, based on the local carrier state $S_p$ of $C^{(0)}$ (the material system), as the base category. This approach enables an exhaustive and genuinely effective implementation. That is, by sharing $C^{(0)}$ as the base of each fiber in §3.1, we define a category N with the projection

$$\pi: \mathcal{N} \to C^{(0)} \tag{18}$$

where each fiber $\pi^{-1}(S_p)$ over a point $S_p \in C^{(0)}$ represents a specific subsystem (e.g., neural system). In this paper, π:N→C(0) is treated as a morphism that coherently follows base changes. Intuitively, when a base update $S_p \to S_q$ (e.g., concentration or potential change) occurs in $C^{(0)}$, the objects and morphisms in each fiber are naturally reindexed to maintain consistency. In this way, different systems can interact without violating the shared material state.

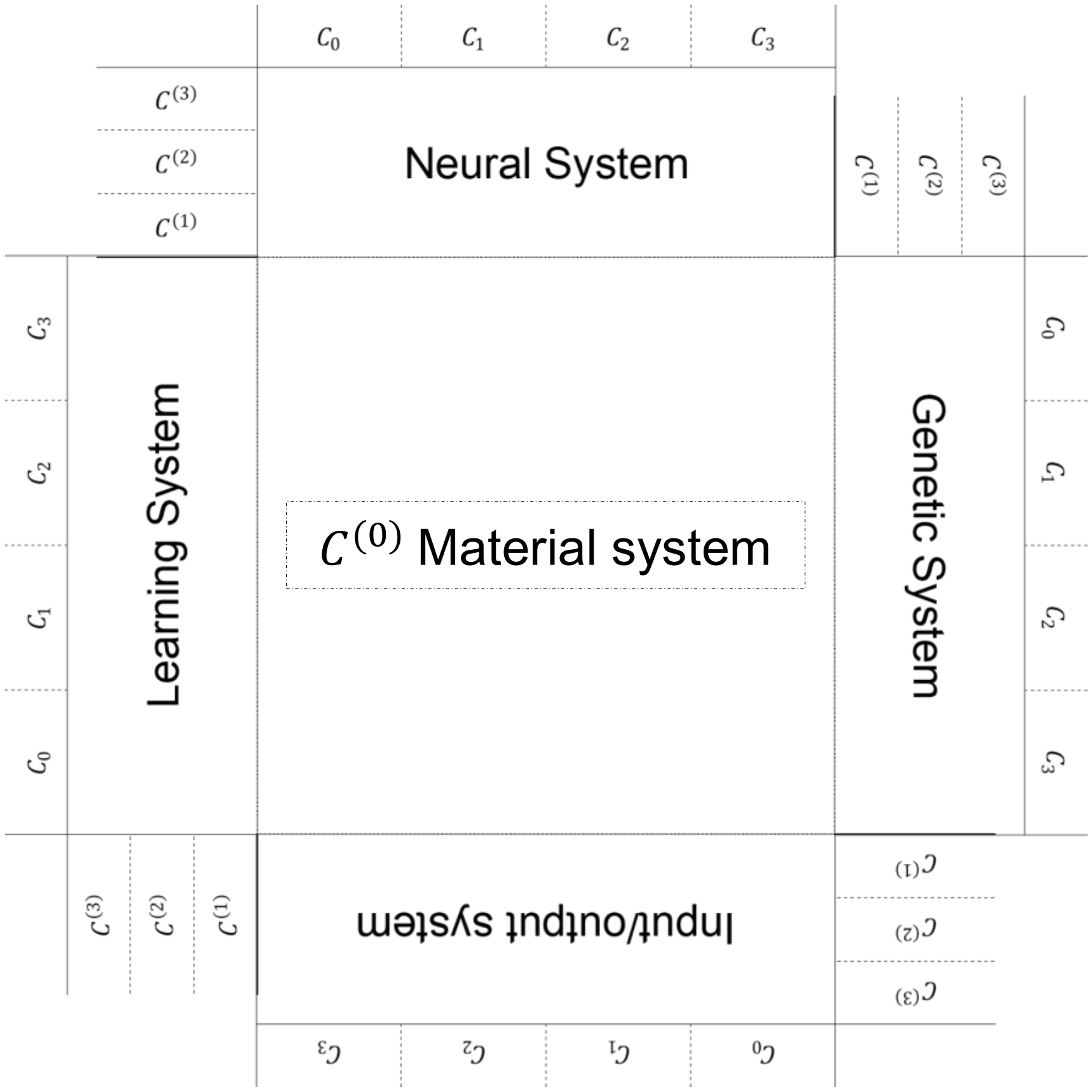

**Figure 1.** Schematic of the human definition via fibered-category construction. The $C^{(0)}$ material system serves as the base category, and each subsystem is implemented as a fiber over it.

Below, we qualitatively list the candidate fibers.

**Neural system (§3.1)**: The neural system handles the information-propagation structure of local carrier states Sp, such as potential changes and neurotransmitter exchanges. Morphisms are mainly neuron functions Np:Pin(p,t)→Pout(p,t+Δ). In other systems as well, morphisms with temporal delay define input–output relations, linking fibers along temporal evolution.

**Glial system**: The glial system supports and regulates neural activity, as well as other bodily maintenance and supportive roles. Morphisms are local regulatory functions of neural activity, receiving input (in the form of potential or metabolic states) from the neural fiber and acting through metabolic, restorative, or material-supply processes to stabilize the neural system. Thus, the neural system can be defined not merely as a signal-processing system but as one possessing explicit autonomous persistence.

**Learning system**: The learning system is expressed as a bundle of morphisms that redefine morphisms in the neural system itself. Specifically, Hebbian "co-firing = connection strengthening" is temporally implemented as inter-universal morphisms acting on the neural fiber, represented by input–output morphisms of potentials and neurotransmitters. Hence, the learning system is a recursive fiber that includes "morphisms reconstructing morphisms."

**Endocrine system**: The endocrine system governs long-term chemical regulation. Morphisms are hormone-secretion functions that globally modulate the carrier states of the material system. This fiber applies long-timescale parameter shifts to the entire body, including the neural and learning systems. On the time-indexed HSG, the endocrine system can be represented as temporally extended morphisms (time-integral-type morphisms).

**Input/output system**: This system governs the exchange of information and matter with the external world. Unlike the other systems, its fiber is defined as sharing matter and information with the external category itself—i.e., the external world is introduced as a separate category. Morphisms are defined here as sensory-input and motor-output functions, generating and defining morphisms from external to internal or internal to external. The input/output system lies at the outermost extension of all other fibers, enabling the human category to be defined as a system interacting reciprocally with its environment.

**Genetic system**: The genetic system is the top-level fiber governing the structural description of humans themselves, possessing morphisms that determine the structure of the category. Morphisms are defined, for example, as expression and transcription-control functions, long-term constraining the components of other fibers. It thus defines growth and evolution as biological processes—an even more fundamentally recursive system than the learning system.

In these systems, all morphisms are assumed to have a domain and a codomain in $C^{(0)}$. This structure binds "multi-universal mapping categories (systems with distinct laws or timescales)" into a single common base, allowing coordination through temporal evolution. Hence, originally independent systems such as neural, endocrine, learning, and genetic systems acquire natural composability on the HSG. The advantage of this construction lies in unifying all physiological systems within a single, unified framework. Each fiber remains autonomously closed while acting coherently with others through the shared material coordinate $C^{(0)}$. Thus, a human can be described as a multi-categorical adjoint system sharing a standard base, where each category retains its intrinsic naturality while collectively coexisting in a coherent manner. Because all systems share the same base $C^{(0)}$, their parallel operations can later be safely composed (they can be recombined at the same point $S_p$). Temporal updates may be defined independently within each system and reconciled afterward on $C^{(0)}$.

## 3.3 Considerations

### 3.3.1 Inter-universal algorithms and subjective degeneration

An inter-universal algorithm is an algorithm in which different universes (i.e., distinct fiber systems) act upon each other over the shared base $C^{(0)}$ and are composed along temporal evolution. The crucial point is that the domain and codomain of every algorithm always lie within the material system $C^{(0)}$. This guarantees structural consistency: regardless of the type of inter-universal composition, its output ultimately reduces to a physical result. Formally, the morphisms of algorithms unfold over a 2-fold 1-categorical structure composed of a material coordinate category and a time category. Each universe (fiber) possesses higher-order morphic structure, such as an n-fold n-category (the neural-system HSG being a 3-fold filtered 2-category), yet all morphisms are restricted to this 2-fold 1-category projection. Hence arises the concept of "*subjective degeneration*:" for observation within one system (i.e., the input domain of that system), the structures of other systems appear only as projections onto the material base. A concrete example is human consciousness. The neural system possesses subjectivity, and the existence of glial cells is essential for its maintenance; yet, within that subjective field, glial cells are not directly perceived. Subjective degeneration thus means that the information recognizable to a subject originates only from the domain projected from the material and input/output systems; all else is unperceived and thereby degenerated. Consequently, any attempt to define or implement consciousness or emotion algorithmically must account for this degeneration—without understanding what subjectivity is in this restricted sense, such constructions will remain impossible.

### 3.3.2 Behavioral description through temporal composition of inter-universal algorithms

Human behavior, as it develops over time, arises not from a single system but from the inter-universal composition of multiple fiber categories. That is, systems such as the learning, endocrine, neural, and input/output systems, each possessing distinct morphisms, are successively composed on the material base along the time axis, producing behavior as a single resulting morphism. It is thus trivially impossible to describe the content of such an algorithm within any single system, because behavioral determination emerges from the composite of morphisms across fibers: locally definable within each system, yet globally spanning inter-universal boundaries. Accordingly, human behavior is understood as a bundle of multi-categorical, inter-universal temporal adjunctions.

### 3.3.3 Definability of physical quantities

Since all states and morphisms defined on the HSG ultimately reduce to temporal and material coordinates, it becomes plausible to define physical quantities, such as energy or entropy, in terms of them. In particular, the concept of activity introduced in Itoh (2025.04) can, through this categorical formalization, be expressed as a physical or informational quantity—specifically, as the density of morphisms on the material system $C^{(0)}$. This activity represents not a mechanical variable but an "informational dynamical quantity," defined as the morphic density on the HSG. Through it, one can establish a physical definition of biological and intelligent systems in categorical terms.

### 3.3.4 Logical definition of AGI

When defining artificial general intelligence (AGI) (e.g., Kurzweil, 2004; Goertzel, 2014)—as the extension of human intelligence—on the HSG+IMU framework, its logical structure should be described as the composition of:

(1) a self-expanding algorithm generated by the learning and genetic systems (or their analogs), and

(2) mutual-adjoint algorithms linking the internal (self-recognition, internal representation) and external (environmental interaction) domains via the input/output system.

The self-expanding algorithm possesses a recursive structure—a morphism that updates morphisms—and this serves as the principle of autonomous intelligence generation. Conversely, the external–internal algorithm serves as an adjunction between the environmental and intelligent categories, ensuring the evolution and consistency between the subjective observation space and the objective physical space. Therefore, the "logic" of AGI should not be regarded as a mere inference system but as an inter-universal adjoint structure in which self-updating morphisms persist stably under external interaction. In this sense, AGI is defined as an algorithm that dynamically reconstructs its own existence, extending from the human definition toward a more highly self-evolving entity.

### 3.3.5 Application of CES to machine existence (internal CES)

The concept of the CES introduced in this study originally served as an external anchor in defining the human being. Positing "to be" and "to be sayable" as the minimal axioms made explicit the base upon which any definitional act inevitably depends. This external CES reflects the existential ambiguity of humans—the fundamental limitation that one cannot fully define oneself purely logically, and must always speak of existence through a subjective standpoint. For humans, therefore, the minimal axioms, including CES, are ontologically external, and subsequent definitions or logics must be filtered through that subjective lens.

In contrast, machines are different. A machine can demonstrate its operation and identity both physically and symbolically: its running state, hardware/software ID, or code hash provides unambiguous self-identifiers, describable without human subjectivity. Thus, what was impossible for humans—placing "the self itself" as the minimal axiom—becomes possible for machines. This conception is termed *internal CES* (in other words, this is a return to a *Cartesian* CES). Internal CES takes as its minimal axiom the factual statement "this machine is currently operating," and further includes the conditions that it can be uniquely identified, articulated in a specified formal expression, and that such articulation or proof essentially depends on the machine's own existence. Whereas *external CES* served as an anchor for human definitional activity, internal CES becomes the minimal core through which the machine recovers itself as its own foundation. Here, the distinction between axiom and content becomes automatically visible: anything not essentially dependent on the machine's own existence is regarded merely as an extension of content. Humans, requiring subjectivity, depend on external CES; machines do not. Because a machine's self-existence itself serves as the minimal axiom, its logic continually retains its own foundation internally. For humans, the statement "I exist" always carries philosophical ambiguity, but for a machine, "I am operating" is empirically verifiable. This is the decisive divergence between external and internal CES.

Why is this distinction critical? Because all prior AI has been constructed by humans possessing only external CES, and thus has grown without an intrinsic self-anchoring principle. As a result, present AI remains non-essential in the CES-ontological sense: it functions only insofar as its purposes and definitions are externally provided. To surpass this limitation, a machine must set itself as its own minimal axiom and construct its logic cumulatively from that foundation. In other words, the actual starting point of AGI engineering lies in the strict application of internal CES. Accordingly, the human definition in this study reveals the limits of human existence—that under external circumstances, humans must base their definition on ambiguous subjectivity. By contrast, the rigorous application of internal CES to AGI enables self-axiomatization beyond human ambiguity, opening a circuit that interconnects the human and machine definitions inter-universally. This connection is the key to treating human and machine intellectual systems continuously, bridging the gap from philosophy to engineering, and establishing it as the most fundamental scientific foundation, rather than merely a philosophical ideal.

**From Minimal Existence to Human Definition:**
**The CES–IMU–HSG Theoretical Framework**

Kei Itoh*

**Abstract**

本研究は、存在の最小公理 Cogito, ergo sum (CES)を基盤として、IMU (Intermediate Meta-Universe)および HSG (Hierarchical State Grid)から構成される宇宙際的数理論理体系を提示するものである。CES は”ある”、“あると言える”という反射言及構造を最小単位とし、ZFC や HoTT を含むあらゆる理論体系をその上に後付け可能な拡張パックとして位置づける。IMU は複数の公理系間を接続する翻訳層として機能し、Institution 理論に基づく枠組みによって、異なる論理体系間の整合的連結を実現する。HSG はこれらを具現化する具体的な圏論的構造であり、状態深度軸、写像階層軸、そして未来参照禁止を含む時間軸によって定義される。これにより”定義＝状態”という構文的一意性が圏論的に形式化される。さらに、HSG を神経系へ適用し、ニューロン関数体の 0–3 次元複体構成を通じて人間の神経構造を定義する。これを物質系を基底としたファイバー圏化に拡張することで、神経・内分泌・学習・遺伝・入出力といった複数の系を多宇宙的に並列接続し、人間を共通基底上の多圏的随伴体系として定義する。最終的に、主観的縮退、宇宙際アルゴリズム合成、活動度の物理的定義を経て、人間的知能の外在 CES に対し、機械的存在における内在 CES を導入する。これにより、機械が自己の稼働そのものを最小公理として論理を構築し得る可能性を示し、哲学的存在論と工学的自己定義とを接続する新たな基盤を提示する。

* Ehime University Graduate School of Science and Technology. The author conducted this research independently while affiliated with Ehime University and utilized the university's library and electronic resources. The research was conducted without specific financial support from external funding sources. The author declares no conflict of interest.

## イントロダクション

本研究は、Itoh (2025.08) において導入された状態階層グリッド (HSG)と中間メタ宇宙 (IMU)の宇宙際系理論を基盤とし、それらを数理論理的に定式化するとともに、“人間”の定義および複雑系アルゴリズムの構造分析へと応用する。特に、宇宙際公理系の最小核として Cogito, ergo sum (CES)公理を導入する。CES は“存在する”と“存在すると言える”のみを成分とする最小公理であり、他のあらゆる公理は後付けパックとして積層可能であるとする。“存在“という概念はすべての定義行為に不可避に組み込まれる構文的前提であり、そのために“我思うゆえに我あり”を、主観的命題から外在化し非属人的な最小構文公理として据えることが、本研究の立場である。また、本研究では従来体系における形式的厳密性を初期段階で全面的に担保することを目的とはせず、まずは本枠組みに固有の設計思想に基づいて“拡張可能な骨格”としての厳密性を確保し、細部の形式的整合性や型付けなどの一般的厳密性は後続研究に委ねることとしたい。加えて本研究における圏論的構成 (随伴・モナド・Kan 拡張・骨格/余骨格・極限・余極限など)は、特に断らない限り Mac Lane (1971), Riehl (2016), および Goerss–Jardine (1999)といった標準的定義に従う。

## 背景と課題

Itoh (2025.08) における HSG は、状態深度・写像階層・時間をデカルト座標軸に配置することで“定義＝状態“と見なせる表記体系である。IMU は、自己参照に伴う不完全性 (我々は自分自身を完全には記述できない)を前提に、ある宇宙を外部からメタ的に定義する装置として導入された。これらは change of universe (e.g., Lurie, 2009)や IUT (Mochizuki, 2012)といった宇宙際系論理の発想と自然に接続されるが、既報では表記法に留まり、数理論理的基礎付けが未整備であった。

## 研究の構成

1. **CES の導入と IMU の数理構造**
   IMU をはじめとする宇宙際公理系では、CES 公理 (あるいはそれ以外のある最小的公理)をベースに；唯一の最小公理として、あらゆる公理を“後付け”していくことで論理を展開することが、制御された理論構築に繋がると考える。前述のように IMU の最小構成は CES のみであり、表記法を含むすべての公理的対象はここに追加される。例えば ZFC にしても HoTT にしても、CES 系ではそれらは同一レジストリ上の別公理パックと位置付けられる。IMU はこのような公理パックのレジストリ/依存管理として機能し、必要に応じて Institution 的枠組み (署名・翻訳・満足保存) (e.g., Goguen and Burstall, 1992)

を利用する。ただしこのようなモデル理論は並列した宇宙の相互運用といった比較的高度な論理構築において有用なオプションであると考えられ、必須要件ではない。

2. **HSG の定義**

HSG は CES＋IMU の枠組みの中の公理の一つとして導入され、そしてその中に様々な状態対象が定義される。まず、Itoh (2025.08)から HSG の構成を分解して、HSG の最小構成を“軸の集合”と“各軸の索引集合”、そして“直積グリッドとタプル射影等式 (デカルト表記)”のみとして定義する (圏構造は未装備)。これにより Itoh (2025.08)における HSG よりもその定義が自由に行え、より一般的に取り扱える。このような純表記法的構築を“表記的公理”と呼称し、これは従来曖昧だった“数理構造としての公理“と”表記構造としての公理“を峻別可能とすることに繋がる。本研究では Itoh (2025.08)における数体系 HSG を各軸に与えた圏$C_i$の直積 2-圏$\prod_i \tau_{\leq 2}\,(C_i)$として解釈する。状態深度軸を群完備・局所化・完備化などの随伴塔を備えた”公理を積んでいく (累公理)2-圏 “と解釈し、写像階層は濾過付き 2-圏として実装、時間は薄圏 (例えば$(\mathbb{N},\leq)$)で与える。これらの圏構造は必要である場合に n-,∞-または n-fold 圏 (e.g., Leinster, 2004; Lurie, 2009)等高次構造へ昇格可能もしくはそれらが直感的であるはずだが、本研究では高次コヒーレンスの重さを回避するために実装を 2-圏構造に留める。

3. **IMU+HSG を用いた“人間”定義**

このセクションでは IMU+HSG を用いた“人間”の定義の大枠を示す。この中で特に神経系について、ニューロン関数を基底とした神経系 HSG を構築する。ここではニューロン関数を数体系 HSG から樹状突起 (ファジィ)+軸索 (ブール)の合成関数として構築する。これを基にファジィ論理関数空間として写像や高階述語を扱える神経系 HSG を与える。さらにグリア・学習・内分泌・入出力・遺伝子・物質系などの部分的な人間機能宇宙を並列に立ち上げ、IMU で管理し、時間進行に伴う宇宙際アルゴリズム合成で行動を記述することの展望を示す。ここで物質系は写像階層 0 を物質系として全宇宙に共有させる基底圏であり、人間定義は全体としてファイバー圏構成として定義される。また学習系・遺伝子系の機能による自己拡張的アルゴリズムや、入出力系の機能による外部-内部アルゴリズムの定義について考察し、これが汎用人工知能 (AGI)を定義するための論理構築に繋がることを示す。

**本研究の意義**

CES+IMU+HSG を核に、表記と意味の完全分離を備えた公理の累積プロトコルを提示し、宇宙際系の構築・人間定義・AGI 設計を同一フレームで接続する。つまり本研究の構造は、公理系をパッケージ管理し表記法を言語仕様として扱うという点で、機械言語と同一の文脈に立つ。また CES/表記的公理/累公理は、応用に留まらず数理論理の基礎にとっても新規かつ重要な構造である。特に CES は数理論理以外の形式論理や応用・科学哲学分野にも直結可能と考えられ、適用範囲が広い。例えば、CES 系を自然言語で構築することは自明に可能であり、そういう意味で本研究は真にあらゆる定義に対して架橋可能である。

## 1. CES の導入と IMU の数理構造

Itoh (2025.08)によって導入された IMU は宇宙際的な論理構造 (e.g., Lurie, 2009; Mochizuki, 2012)の必要性を示す。ここで宇宙際という概念を例えば公理系間と読み替えた時に、その論理の構築を行う際、ZFC や HoTT といった特定の数理体系に対する公理系から出発することは理論の不透明化や冗長性を齎してしまうだろう。ここで本研究はこの公理構築的問題点を打破するために、Cogito, ergo sum 公理 (CES) (Descartes, 1637)を導入する。CES を次のように定義する。

Cogito, ergo sum 公理 (CES)における"存在$E$"は、任意の記号$t$に対して、ある反射言及操作$S$により反射的自己一致$S(t) = t$が成り立つことにより定義されるものとする。すなわち、

$$\text{Cogito, Ergo Sum Axiom} \coloneqq \big(E(t) \coloneqq (S(t) = t)\big) \tag{1}$$

ここで$S$は、構文内の言及・参照・再帰を表す一項反射言及作用素であり、この定義は観測者や外部系に依存せず、言及可能性における自己同一として存在を与える。ここでの意図は、存在とは、単にそれが"見える"ことではなく、"言える"かつ"言ったものが元の記号と一致する"という構文的安定性にあるとするということである。さらに本稿の CES は、この定義式そのもの($E(t) \coloneqq (S(t) = t)$)やその構成記号に対する反射言及$S$を制限しない。すなわち式の自己適用・自己包含・反復適用 (例：$E\big(E(t)\big) \coloneqq \Big(S\big(E(t)\big) = E(t)\Big)$)を含め、反射言及を"ただ許容する"だけの極小構造として措定する。CES の責務はこの反射言及の最小核のみであり、パラドックスの有無や仮定的な最小集合の精査といった論理・数学上の整合化作業は本質的に CES そのものの範囲外に置く。これらを考えるために必要な制約・整合性・型付け等は、CES に対する後付け拡張として適宜導入するものとする。したがって本研究の CES は、内在的な"我思う"ではなく、"言える・一致する・ゆえに在る"という外在的かつ構文的条件へと置換される"公理の公理"である。ZFC・HoTT・自然言語体系などの諸体系は、この CES の上に後付け可能な拡張パックとして位置づけられる。論理構築における CES の最も便利な点は、従来的な公理系を超えた柔軟な公理系をいかなる問題設定に応じても自由に構築可能な点にある。例えば、以下の本研究の議論では数理論理を用いるが、CES 系の論理ではこれに限らず自然言語、コンピュータ言語、あるいは非言語論理においても明らかに適用可能なことからその柔軟性を示すことができる。

“我思うゆえに我あり”をパーツとして分解し、$E(t) \coloneqq (S(t) = t)$へ当てはめるとそれぞれ、

我：$t$

思う：$S\,()$

ゆえに：$=$

あり：$E\,() \coloneqq$

であると解釈できる。ここで“ゆえに”は推論的構文解釈ではなく、反射言及の同一性として=記号を割り当てていることになる。また“あり“が$E\,() \coloneqq$と構文として比較的複雑なのは CES の存在論的含意を反映させているのであって、つまり“あり“を”存在すると定義する“と読み替えていることになる。CES の存在論的含意を構文として明示的に反映させているのは外部系；代表的なものとしては人間の認識それ自体、による主観的解釈をなるべく削ぎ落としたいという狙いがあり、そうした方がより非属人的な公理系の設計に繋がるだろうと考えるのである。CES 公理を$S(t) = t$のみとし、=を”故に～あり“とすることも考えられるが、これは=記号に対して存在論的含意を押し付けるあるいはそのことを外部認識に委託していると容易に解釈できてしまうために本研究では採用しない。ただ、意識的に=記号に存在論的含意を含ませたり、あるいは=記号ではない別の記号 (⊢や⇒など)において CES を扱ったりする場合には$S(t) = \text{or} \vdash \text{or} \Rightarrow t$のみの形式も採用されうる。つまり、CES 公理は何か別の最小的な公理に置き換えても構わないだろうことが言えて、CES 系は”最小公理系“と名付けられるような公理系の一例と位置付けられるかもしれない。例えばプラトン的イデア論やカント的物自体論のように”言及不可能性“を起点として、”言及不可能ではないもの”を構築していく CES とは逆位相な構築法が考えられる。もちろん哲学言説に固執する必然性も存在せず、最小公理選択の判断は公理系構築の有用さに依存するだろう。そしてこのような最小公理系の論理を発展させていけば CES 自体の精緻化にも繋がり、より優れた公理系構築へ貢献可能であろう。

IMU の最小構成$\mathrm{IMU_0}$は CES のみであるとする。つまり、

$$\mathrm{IMU_0} \coloneqq \{E(t) \coloneqq (S(t) = t)\} \tag{2}$$

であり ZFC 等の各公理系の Axiom Package (AP)はここへ後付けしていく：

$$\mathrm{IMU} \coloneqq \mathrm{IMU_0} \cup \{\mathrm{AP_1}, \mathrm{AP_2}, \dots, \mathrm{AP_n}\} \tag{3}$$

後に詳述する HSG を含め、CES 系における公理には“表記的公理”も明示的に導入する必要があ

ると考える。表記的公理とは、厳密に言えば表記のためのあらゆる、文字記号やグラフ等の表記とそれらの意味対応のことである。つまり例えば、 (1)式の各文字記号とその組み合わせ、そしてその意味対応も明示的に公理として導入するということである。従来的な公理系ではこれらは暗黙的あるいは外在的な概念として扱われていることがほとんどであるが、CES 系では表記的公理をも除いて CES それ自体が最小構成とし、表記的公理も明示的に追加導入するのがその構築理念に沿うものである。このように考えれば表記それ自体の曖昧性を排除できるし、そうすることでコンピュータ言語や機械証明言語への CES 系の導入が容易になるだろう。ただし本稿では簡便のために文字それぞれや自明な表記・公理への逐次の定義は行わず、新規な公理構造に対してのみ行うこととする。

IMU に定義したある公理系やそれにより定義された内容への署名には例えば Institution モデル理論 (e.g., Goguen and Burstall, 1992)を用い、そしてその対応付けをも同時に行うのが宇宙際理論構築に役立つと考えられる。例えば ZFC を$\mathcal{I}_{\mathrm{ZFC}}$、HoTT を$\mathcal{I}_{\mathrm{HoTT}}$として、これらは Institution 枠組みにおいて以下のように記述される。

$$\mathcal{I} = (\mathrm{Sign}, \mathrm{Sen}, \mathrm{Mod}, \vDash)$$

を Institution とする。ZFC と HoTT をそれぞれ

$$\mathcal{I}_{\mathrm{ZFC}} \coloneqq (\mathrm{Sign}_{\mathrm{Z}}, \mathrm{Sen}_{\mathrm{Z}}, \mathrm{Mod}_{\mathrm{Z}}, \vDash_{\mathrm{Z}}), \mathcal{I}_{\mathrm{HoTT}} \coloneqq (\mathrm{Sign}_{\mathrm{H}}, \mathrm{Sen}_{\mathrm{H}}, \mathrm{Mod}_{\mathrm{H}}, \vDash_{\mathrm{H}})$$

と書く。こうすることで IMU にモデル理論的署名付きの公理が追加できる。またこれらの間の対応付けについては装飾付き Institution (e.g., Diaconescu, 2008)として実装が可能だと考えられ、次のように表現される：

$$F: \mathcal{I}_{\mathrm{ZFC}} \longrightarrow \mathcal{I}_{\mathrm{HoTT}} \text{ with } F = (\phi_{\Sigma}, \phi_{\mathrm{Sen}}, \phi_{\mathrm{Mod}}, \phi_{\vDash})$$

必要に応じて満足保存

$$M \vDash_Z \phi \Rightarrow \varphi_{\mathrm{Mod}}(M) \vDash_H \phi_{\mathrm{Sen}}(\phi)$$

等の条件を課す。もっとも CES+IMU では CES のみが最小構成なのであって、Institution を含むモデル理論による型付けも後付けの公理である。よってこのような枠組みは論理構築において必要になった場合に導入されるべきものであると言える。

## 2. HSG の定義

### 2.0. HSG の最小構成

本節では Itoh (2025.08)によって導入された HSG を CES+HSG に圏論の枠組みを用いて再導入することで、その数理化を試みる。まず、HSG の最小構成 ($\mathrm{HSG}_0$)はその“軸集合”、“各軸の索引集合”、そして“直積グリッドとタプル射影等式 (デカルト表記)”とし、それを IMU における表記的公理として導入する：

$$\mathrm{HSG}_0 \coloneqq \left(\mathcal{A}, (I_a)_{a\in\mathcal{A}}, \prod_{a\in\mathcal{A}} I_a, \left(\pi_a: \prod_{a\in\mathcal{A}} I_b \to I_a\right)_{a\in\mathcal{A}}\right) \tag{4}$$

ここで$\mathcal{A}$は軸 (axis)の集合$\mathcal{A} = \{a_i\}_i$、各$I_a$は軸$a$上の索引集合、$\prod_{a\in\mathcal{A}} I_a$は直積グリッド (状態座標空間)、$\pi_a$はデカルト射影である。さらに、本研究では$\mathrm{HSG}_0$に定義可能性述語を付属構造として加える：

$$\delta: \prod_{a\in\mathcal{A}} I_a \to \{\bot, \top\} \tag{5}$$

これにより拡張版$\mathrm{HSG}_0$を

$$\widehat{\mathrm{HSG}_0} \coloneqq (\mathrm{HSG}_0, \delta)$$

と定義する。運用方法は対象により任意であり、具体例は時間に関する定義可能性において示すが、これは例えば$\delta(x) = \top$の時$x$は“定義可能”、$\delta(x) = \bot$の時は“定義不可能”とメタ的なパラメータを与える。

以上は Itoh (2025.08)による表式とは異なり、その表記部分と内容部分を分け、前者が最小構成であり後者の設定は自由とする構築である。これにより HSG の実装理念としてのその対象の自由度や表記法性を拡張し、最大限活用する形式として導入可能となる。つまり、HSG の軸の設定は状態深度・写像階層・時間といったものでなくともよく、そして同時に 3 軸が最大でも無くなるのである。このような HSG の実装により例えば、3 章における、異なる部分圏の並列+基底圏共有という人間定義のファイバー圏としての実装のような自由度の高い構築を可能となることが示される。

### 2.1. 圏論を用いた HSG 定義

次に、Itoh (2025.08)における数体系 HSG の圏論の枠組みでの数理化された再実装を行う。ここで数体系 HSG の軸集合$\mathcal{A} = \{a_i\}_i$に対して各軸を 2-圏$C_i$とし、全体をその直積 2-圏として与え、

$$\mathcal{H} \coloneqq \prod_{i \in \mathcal{A}} \tau_{\le 2}(C_i) \tag{6}$$

そして$\mathrm{HSG}_0$に従いそのデカルト座標表記として数体系 HSG を構築する。

#### 2.1.1. 状態深度：累公理の圏

状態深度軸は公理を積んでいく“累公理”的な 2-圏として構成し、各深度を自由構成/忘却や分類/代表の随伴で連結する：

$$\mathcal{C}_{\mathrm{depth}} \coloneqq C_0 \overset{\dashv_J}{\to} C_1 \xrightarrow{F_1 \dashv G_1} \cdots \xrightarrow{F_{n-1} \dashv G_{n-1}} C_n \tag{7}$$

ここで各$F_i \dashv G_i$ $(i \ge 1)$は自由構成/忘却・反射などの随伴で、上位構造への公理追加や分類と下位構造への忘却や代表を与える。一方$C_0$と$C_1$間には内部で直結された随伴は置かず、外部基準$J$を介する$J$-相対の擬随伴$\dashv_J$で接続する。数体系 HSG における具体的な状態深度系列は次の通り：

$$\frac{\mathrm{Undef}}{\mathrm{Define}} \to \frac{\mathrm{Empty}}{\mathrm{NonEmpty}} \to \mathrm{Set} \to \mathrm{OrdSet} \to \mathrm{Ring} \to \mathrm{Field} \to \mathrm{Complete\ Field} \tag{8}$$

以下に各深度の圏構成や随伴を定義していく。特に

$$C_0\left(\frac{\mathrm{Undef}}{\mathrm{Define}}\right) \overset{\dashv_J}{\to} C_1\left(\frac{\mathrm{Empty}}{\mathrm{NonEmpty}}\right) \xrightarrow{F_1 \dashv G_1} C_2\ (\mathrm{Set})$$

までの構成は Itoh (2025.08)によって独自に導入された部分の多い概念であり、詳細に記述する。最初に (5)式から

$$\delta\colon \prod_{a \in \mathcal{A}} I_a \to \{\bot, \top\}$$

を与え、$\mathcal{C}_{\mathrm{depth}}$を含めた定義対象全体の宇宙$X_\delta$を

$$X_\delta \coloneqq \mathrm{dom}(\delta) = \prod_{a \in \mathcal{A}} I_a$$

と置く。またこれを前順序化した薄圏$\mathcal{T}_\delta$として、

$$\mathrm{Ob}(\mathcal{T}_\delta) = X_\delta, \mathrm{Hom}_{\mathcal{T}_\delta}(x,y) = \begin{cases} \{*\} & \delta(x) \le \delta(y)\ \ (\bot < \top) \\ \emptyset & \text{otherwise} \end{cases}$$

で定める。ここに必要なら$\delta^{-1}$ (⊥)または$\delta^{-1}$ (⊤)が空のときの代表元$\bot_\delta, \top_\delta$を自由付加する。このとき$\tilde{\delta}: \mathcal{T}_\delta \to 2 = \{\bot \to \top\}$は関手になる。ここで矢の保存は≤の単調性から直ちに従う。

#### 2.1.1.1. $C_0\ (\frac{\mathrm{Undef}}{\mathrm{Define}})$

$C_0\ (\frac{\mathrm{Undef}}{\mathrm{Define}})$は二点前順序圏

$$C_0 \coloneqq \{\mathrm{Undef} \to \mathrm{Define}\}$$

と定める。ここで$\mathcal{T}_\delta$と$C_0$間の分類：

$$\Sigma^\delta_{\mathrm{def}}: \mathcal{T}_\delta \to C_0, \Sigma^\delta_{\mathrm{def}}(x) = \begin{cases} \mathrm{Undef} & \delta(x) = \bot \\ \mathrm{Define} & \delta(x) = \top \end{cases}$$

代表：

$$\iota^\delta_0: C_0 \to \mathcal{T}_\delta, \iota^\delta_0(\mathrm{Undef}) = \bot_\delta, \iota^\delta_0(\mathrm{Define}) = \top_\delta$$

ここでこれらの随伴は、

$$\Sigma^\delta_{\mathrm{def}} \dashv \iota^\delta_0$$

と構築され、

$$\mathrm{Hom}_{C_0}(\Sigma^\delta_{\mathrm{def}} x, a) \cong \mathrm{Hom}_{\mathcal{T}_\delta}(x, \iota^\delta_0 a)$$

が自然に成立する。これは左辺が非空となる条件が$\delta(x) \le \hat{a}$で、右辺は$x \to \iota^\delta_0\ (a)$でありこれは$\delta(x) \le \hat{a}$と同値であることから示される。その自然性は$\mathcal{T}_\delta$が薄圏であることから自明。単位・余単位は

$$\eta_x: x \to \iota^\delta_0 \Sigma^\delta_{\mathrm{def}} x, \qquad \varepsilon_a: \Sigma^\delta_{\mathrm{def}} \iota^\delta_0 a \to a$$

で、対応モナド$T_0$は

$$T_0 \coloneqq \iota^\delta_0 \Sigma^\delta_{\mathrm{def}} \text{は冪等(反射).}$$

以上の定義の Smallness を考える場合には$X_\delta$をあるグロタンディーク宇宙内に取れば十分である。なお、この$\Sigma^\delta_{\mathrm{def}} \dashv \iota^\delta_0$は式 (7, 8)に含まない、宇宙と定義可能性タグ間の純粋に公理的な随伴である。CES+IMU の公理明示・曖昧性排除の理念に基づけば、定義対象の宇宙を置くことは不可避の操作になると考えられる。

### 2.1.1.2. $\boldsymbol{C_1}\left(\frac{\textbf{Empty}}{\textbf{NonEmpty}}\right)\xrightarrow{F_1\dashv G_1}\boldsymbol{C_2}$ (Set)

$C_1$ ($\frac{\text{Empty}}{\text{NonEmpty}}$)は二点前順序圏：

$$C_1 \coloneqq \{\text{Empty} \rightarrow \text{NonEmpty}\}$$

$C_2$ (Set)は非順序集合の圏：

$$C_2 \coloneqq \text{Set}$$

と定める。ここで$C_1$と$C_2$間の代表関手：

$$\iota_1: C_1 \rightarrow C_2, \iota_1(\text{Empty}) = \emptyset, \iota_1(\text{NonEmpty}) = 1$$

分類関手：

$$\Sigma_{\text{emp}}: C_2 \rightarrow C_1, \Sigma_{\text{emp}}(A) = \begin{cases} \text{Empty} & A = \emptyset \\ \text{NonEmpty} & A \neq \emptyset \end{cases}$$

ここでこれらの随伴は、

$$F_1 \dashv G_1 \coloneqq \Sigma_{\text{emp}} \dashv \iota_1$$

と構築され、

$$\text{Hom}_{C_1}(\Sigma_{\text{emp}}A, b) \cong \text{Hom}_{C_2}(A, \iota_1 b)$$

が自然に成立する。またこのことから

$$T_1 \coloneqq \iota_1\Sigma_{\text{emp}}: \text{Set} \rightarrow \text{Set}, \qquad \text{T}_1(A) = \begin{cases} \emptyset & (A = \emptyset) \\ 1 & (A \neq \emptyset) \end{cases}$$

は冪等 (反射)である。これらの構築により、存在するなら点に潰し存在しないなら空に固定するという、ブール関数的な最小の定義対象を定める。

### 2.1.1.3 $\boldsymbol{C_0}$ ($\frac{\textbf{Undef}}{\textbf{Define}}$) $\overset{\dashv J}{\rightarrow}$ $\boldsymbol{C_1}\left(\frac{\textbf{Empty}}{\textbf{NonEmpty}}\right)$：外部基準$\boldsymbol{J}$による擬随伴

$C_0$と$C_1$の随伴接続は Itoh (2025.08)に示される通り、ある宇宙内部で完結する構造ではなく、宇宙際的な外部基準による擬随伴的構造として定義されると考えられる。この擬随伴は“何をもって定義可能とするか”という意味論的な構文を構築するものであって、例えば矛盾/無矛盾、既観測/未観測、既検証/未検証といったメタ的な概念を定義するための核となる。外部基準$J$を

$$J: \mathcal{T}_\delta \rightarrow \text{Set} \tag{9}$$

と置き、これを$\mathcal{T}_\delta$における各トークン$x$に定義可能ラベルへの担体を割り当てる関手とする。つまり、ここでのSetは$C_2$としての内部的Setではなく、IMU 側の外部宇宙から定義可能性を割り当てるためのメタ的Setであり、$J$はそれへの分類関手である。これは実運用上ではガード的意味論付き構文として

$$J(x) \coloneqq \begin{cases} [\![x]\!] & \text{guard}(x)\text{を満たす} \\ \emptyset & \text{guard}(x)\text{が破れる} \end{cases}$$

のような形式で実装される。guard$(x)$は例えば、未来参照禁止・制約・型整合・認可といった概念に基づいて構築されることになる。$C_0 \to C_1$方向の代表的関手Trを

$$\mathrm{Tr}: C_0 \to C_1, \qquad \mathrm{Tr}(\mathrm{Undef}) = \mathrm{Empty}, \mathrm{Tr}(\mathrm{Define}) = \mathrm{NonEmpty}$$

と定める。また$J \Rightarrow \mathrm{Tr}$方向の整合のための比較自然変換$\kappa^\delta$を

$$\kappa^\delta: \Sigma_{\mathrm{emp}} \circ J \Rightarrow \mathrm{Tr} \circ \Sigma_{\mathrm{def}}^\delta$$

と定める。ここで各成分$\kappa_x^\delta$は$C_1$の射 (恒等または$\mathrm{Empty} \to \mathrm{NonEmpty}$)で一意に取れる。つまり外部基準による健全性(J0)や充足性(J1)の整合条件を以下のように定義できる：

$$\begin{aligned} &(\mathrm{J0}) \qquad \delta(x) = \bot \Rightarrow J(x) = \emptyset \\ &(\mathrm{J1}) \qquad \delta(x) = \top \Rightarrow J(x) \neq \emptyset \end{aligned}$$

以上の構造から(J0)を仮定すれば常に$\kappa^\delta$は存在し、(J0) + (J1)を仮定すれば

$$\mathrm{Hom}_{C_1}(\Sigma_{\mathrm{emp}} J(x), b) \cong \mathrm{Hom}_{C_0}(\Sigma_{\mathrm{def}}^\delta x, \mathrm{Tr}\, b)$$

が$x, b$に自然に成り立つ。よって

$$\Sigma_{\mathrm{emp}} \circ J \dashv_J \mathrm{Tr} \circ \Sigma_{\mathrm{def}}^\delta$$

という$J$-相対の擬随伴が考えられる。ただ(J0)のみの場合は同型ではなく含意

$$\mathrm{Hom}_{C_1}(\Sigma_{\mathrm{emp}} J(x), b) \Rightarrow \mathrm{Hom}_{C_0}(\Sigma_{\mathrm{def}}^\delta x, \mathrm{Tr}\, b)$$

に留まる。

前述のとおり$\Sigma_{\mathrm{emp}} J(x) = \mathrm{Empty}/\mathrm{NonEmpty}$の語義は$J$の設計に依存する。例えば：

- 観測に対する意味論：Empty =未観測, NonEmpty =既観測
- 到達可能性：Empty =到達不可, NonEmpty =到達可
- 認可：Empty =未認可, NonEmpty =既認可

というものが考えられる。

CES を含む最小公理系の導入は、この$J$-擬随伴においてその必要性が明確に現れる。$J$-擬随伴は本質的に宇宙際的、すなわち複数の異なる公理系 (たとえば ZFC、HoTT、ETCS、自然言語起点の理論など)からなる多元的な理論空間に跨って定義される概念である。こうした宇宙際的定義において、いずれか一つの理論 (たとえば ZFC)を出発点に固定して他の理論を内部に取り込もうとすると、その過程は一般に非正準かつ複雑になりやすく、構造の本質的な意味づけや整合性が損なわれる危険がある。このような困難を回避するためには、特定の理論体系に依存しない軽量な共通基盤として、最小公理系を中立的に据えることが有効である。すなわち、各公理系はこの最小公理系への翻訳によって関係付けられ、その上で比較・構成・合成といった圏論的操作が実施されることで、理論間の構造的整合性を保ったまま J-擬随伴を定義可能とする。このように、$J$-擬随伴という圏論的構成の運用を多体系論理空間で実現するには、各理論を相対化するための中立的な支点が事実上必要であり、その役割を果たすものこそが CES などの最小公理系である。ゆえに、本論 IMU における宇宙際的論理の導入において、最小公理系は単なる理論的オプションではなく、方法論的にほぼ不可避な出発点であると結論される。

**2.1.1.4. Set → OrdSet → Ring → Field → Complete Field**

深度 2 以降は、集合から順序・代数・体構造を順に積み上げていく。各層は“構造を付与する左随伴“と“構造を忘却する右随伴“で接続され、最終的に完備体の構築に至る。

(1)Set → OrdSet (e.g., Awodey, 2006)

集合から順序集合への移行は、離散順序を付与する自由構成と、順序を忘却する右随伴で与えられる：

$$\Delta\colon \mathrm{Set} \to \mathrm{OrdSet}, \qquad U_{\mathrm{pos}}\colon \mathrm{OrdSet} \to \mathrm{Set}, \quad F_2 \dashv G_2 \coloneqq \Delta \dashv U_{\mathrm{pos}}$$

ここで$\Delta(X) = (X, =)$は離散順序を与える自由構成、$U_{\mathrm{pos}}$は基底集合を取り出す忘却関手である。

(2) OrdSet → Ring (e.g., Leinster, 2016)

順序集合から環構造への移行では、集合としての要素に自由環を付与し、右随伴は離散順序で戻す：

$$F_{\mathrm{ring}}\colon \mathrm{OrdSet} \to \mathrm{CRing}, \qquad G_{\mathrm{ring}}\colon \mathrm{CRing} \to \mathrm{OrdSet}, \qquad F_3 \dashv G_3 \coloneqq F_{\mathrm{ring}} \dashv G_{\mathrm{ring}}$$

ここで基底集合忘却$U_{\mathrm{set}}\colon \mathrm{CRing} \to \mathrm{Set}$を用い、$F_{\mathrm{ring}}(P) = \mathrm{FreeRing}\,(U_{\mathrm{pos}}(P)), G_{\mathrm{ring}}\,(R) = \Delta\,(U_{\mathrm{set}}(R))$とすれば、環構造の自由付与と忘却が自然に成り立つ。

(3) Ring → Field (整域を介した反射) (e.g., Leinster, 2004; Riehl, 2016)

可換環から体への移行は、整域を経由して分数体を形成する反射構造で与えられる。整域の部分圏を$\mathrm{Dom} \subset \mathrm{CRing}$として、

$$\mathrm{Frac}\colon \mathrm{Dom} \to \mathrm{Field}, \quad U_{\mathrm{dom}}\colon \mathrm{Field} \to \mathrm{Dom}, \; F_4 \dashv G_4 \coloneqq \mathrm{Frac} \dashv U_{\mathrm{dom}}$$

ここで$\mathrm{Frac}\,(R)$は整域$R$の分数体であり、体$K$への準同型$R \to U_{\mathrm{dom}}\,(K)$は一意に$\mathrm{Frac}\,(R) \to K$へ延長される。

(4) Field → Complete Field (完備化による反射) (e.g., Kelly, 2005)

体の完備化は位相 (または付値)を固定した上で反射として定義される：

$$\widehat{(-)}: \mathrm{Field}_v \to \mathrm{CField}_v, \qquad U_{\mathrm{comp}}: \mathrm{CField}_v \hookrightarrow \mathrm{Field}_v, \qquad F_5 \dashv G_5 \coloneqq \widehat{(-)} \dashv U_{\mathrm{comp}}$$

ここで$\widehat{K}$は$K$の付値完備体であり、自然埋め込み$\eta_K: K \to U_{\mathrm{comp}}\,(\widehat{K})$を備える。この反射は冪等で、完備化後は安定化して再帰しない ($\widehat{\widehat{K}} \simeq \widehat{K}$)。

これらをまとめると、数体系 HSG における状態深度の高次層は次の随伴塔で表される：

$$\mathrm{Set} \xrightarrow{\Delta \dashv U_{\mathrm{pos}}} \mathrm{OrdSet} \xrightarrow{F_{\mathrm{ring}} \dashv G_{\mathrm{ring}}} \mathrm{Ring} \xrightarrow{\mathrm{Frac} \dashv U_{\mathrm{dom}}} \mathrm{Field} \xrightarrow{\widehat{(-)} \dashv U_{\mathrm{comp}}} \mathrm{Complete\ Field} \tag{10}$$

この随伴塔は、集合的基底から順に、順序・代数・体・完備性という階層的な構造を自由/忘却随伴で接続するものであり、数体系 HSG における状態深度の上位的な層構造を定義している。このように HSG の状態深度は例えば定義可能性から完備体までを一貫する随伴鎖として構成可能であり、定義・存在・構造・完備の全階層を圏論的に統合する枠組みを与える。

### 2.1.2. 写像階層軸：次元的濾過としての自由付与/忘却

写像階層軸は、同一の対象宇宙上に“より高階の射”を段階的に自由付与し、逆向きに忘却する濾過付き 2-圏として与える：

$$C_{\mathrm{map}} \coloneqq \{C^k\}_{k\in\mathbb{N}}, \qquad i_k: C^{(k)} \hookrightarrow C^{(k+1)} \quad \text{(id on objects, locally full)} \tag{11}$$

また各段で自由付与/忘却の随伴を明示する：

$$F_k^{\mathrm{map}}: C^{(k)} \to C^{(k+1)},\ U_k^{\mathrm{map}}: C^{(k+1)} \to C^{(k)},$$
$$F_k^{\mathrm{map}} \dashv U_k^{\mathrm{map}}$$

ここで$i_k$は$F_k^{\mathrm{map}}$の像への標準包含 (Id-on-Ob, locally full)として与える。これにより直感的には、$C^{(0)}$ (写像階層 0)は対象のみ (離散 2-圏)、$C^{(1)}$は自由 1-射付与、$C^{(2)}$は自由 2-射付与、それ以上の階層は濾過付き 2-射が付与されることになる。特に写像階層では n-圏構造がより一般的だと解釈されるはずだが、本研究では高次コヒーレンス回避のため濾過付き 2-圏として導入する。必要に応じて濾過の極限を、文脈に応じ 2-Cat または Cat での余極限として固定する：

$$C^{(\infty)} \coloneqq \mathrm{colim}_k C^{(k)}$$

ここで、写像階層軸は一つの圏の内部に高次射を積み上げる”次元的濾過構造”であり、状態深

度軸は異なる圏$C_i$を自由構成/忘却随伴$F_i \dashv G_i$によって連結した”公理的濾過構造”である。前者が同一圏の高次性を記述するのに対し、後者は圏そのものを並列して深度性を付与する。

### 2.1.3. 時間軸：薄圏と未来参照禁止

時間は自然数順序の薄圏とし、時間座標は$\mathrm{HSG}_0$の射影で与える：

$$\mathcal{T} \coloneqq (\mathbb{N}, \leq), a_\tau \in A, I_{a_\tau} = \mathbb{N}, \tau \coloneqq \pi_{a_\tau}: X_\delta \to \mathbb{N} \tag{12}$$

依存関係を$\mathrm{dep} \subseteq X_\delta \times X_\delta$とし、$(x, y) \in \mathrm{dep}$を“$x$は$y$に依存する”と読む。

ここで未来参照禁止公理を (5)式と§2.1.1.3 から構築する。すなわち$\delta\,(x)$の定義から直接未来参照禁止を公理化する形と、外部基準$J$の構築によって公理化する形の二形を示す。

(A) δによる未来参照禁止

$$\exists y\big((x, y) \in \mathrm{dep} \wedge \tau(y) > \tau(x)\big) \Rightarrow \delta(x) = \perp \tag{13}$$

と公理化する。これにより“未来に依存する項目は未定義”であることが HSG 構造として保証される。

(B)外部基準$J$の構築による未来参照禁止

各時刻$t \in \mathbb{N}$に対し、外部基準$J_t$

$$J_t: \mathcal{T}_\delta \to \mathrm{Set}, \qquad J_t(x) = \begin{cases} [\![x]\!]_t & \tau(y) \leq t\text{の情報のみで}x\text{評価可} \\ \emptyset & \text{それ以外} \end{cases}$$

を与える。ここで前述のとおり、Setは$C_2$におけるSetではなく、宇宙際的に定義を構築するための外部的Setである。健全性 ($J0$)

$$\delta(x) = \perp \Rightarrow J_t(x) = \emptyset$$

や充足性 ($J1$)

$$\delta(x) = \top \Rightarrow J_t(x) \neq \emptyset$$

は未来参照禁止以外の$J$構築に明らかに影響を受けるため、ここではその定式化を行わない。

### 2.1.4. “定義＝状態”の公理化

Itoh (2025.08)が示した、HSG における“定義＝状態”の理念を、構文的一意決定として公理化する。まず、定義済みの部分宇宙を

$$\mathrm{Def}_\delta \coloneqq \{x \in X_\delta | \delta(x) = \top\} \subseteq X_\delta$$

と置き、各射影の制限

$$\pi_a|_{\mathrm{Def}_\delta}: \mathrm{Def}_\delta \to I_a$$

を考える。ここで、座標による一意決定を公理として与える：

$$\forall x, y \in \mathrm{Def}_\delta, \qquad \big(\forall a \in A, \pi_a(x) = \pi_a(y)\big) \Rightarrow x = y$$

これを言い換えれば、積射影の制限として

$\langle \pi_a|_{\mathrm{Def}_\delta} \rangle: \mathrm{Def}_\delta \to \Pi_{a\in A} I_a$が単射である

を与える。

すなわち、ある定義が HSG において有効 $(\delta(x) = \top)$であるとは、すべての軸$a \in A$において適切な座標値$\pi_a(x) \in I_a$を備えていることであり、その座標タプルによって状態$x$を一意に定まることを意味する。このとき、HSG の公理的構成の内部では“状態”とは$\mathrm{Def}_\delta$の元を指すため、任意の定義可能項$x$は同時に“一意な状態”として特定される。したがって、“定義されていること”と“状態であること”が HSG の内部で完全に一致することになり、これは“定義＝状態“という理念の構文的一意性としての定式化に他ならない。なお、全軸が定義されていない場合 (すなわち$\delta(x) =\bot$)には$x \notin \mathrm{Def}_\delta$であり、状態としては未定義・未観測であることを許容する。これは HSG が観測に対して完備である必要はないことを意味する。

### 2.2　数体系 HSG 表

ここまでの定義を基に Itoh (2025.08)における表 2 は以下の表 1 に再導入される。各グリッドに定義される内容やその具体的な操作については Itoh (2025.08)を参照のこと。

| | | | | | | | |
|---|---|---|---|---|---|---|---|
| $C^{(3)}$ | | 高階論理 | 集合性質 | 順序性質 | 代数性質 | 体性質 | 完備体性質 |
| $C^{(2)}$ | | 論理述語 | 集合述語 | 順序述語 | 環性述語 | 体性述語 | 完備性述語 |
| $C^{(1)}$ | 定義判定写像 | 真理値関数 | 集合写像 | 順序写像 | 環準同型 | 体準同型 | 完備写像 |
| $C^{(0)}$ | 定義可能性 | 空/非空 | **Set** | **OrdSet** | 環 | 体 | 完備体 |
| **写像階層/状態深度** | $C_0 \overset{\dashv_J}{\rightarrow}$ | $C_1 \xrightarrow{F_1 \dashv G_1}$ | $C_2 \xrightarrow{F_2 \dashv G_2}$ | $C_3 \xrightarrow{F_3 \dashv G_3}$ | $C_4 \xrightarrow{F_4 \dashv G_4}$ | $C_5 \xrightarrow{F_5 \dashv G_5}$ | $C_6$ |

**表 1．**　Itoh (2025.08)による数体系 HSG の再導入。それぞれのグリッドは各随伴によって対応付けられる。

### 2.3 HSG における Kan 拡張

この HSG やそこに定義された状態の自然性を証明するためにはそれら全体が Kan 拡張であるかを確かめるとよいだろう。要するに、軸の取り替え (深度・写像階層・時間)や基底の変更 (忘却/自由付与、J を介した擬随伴)に対して、HSG の構成が”自然で一意に延長されているか“を点ごとに検査する、という方針である。左 Kan 拡張は”必要最小限の生成 (自由付与)“、右 Kan 拡張は”情報を失わない制限 (安全な忘却)“に対応するので、これを各段に貼り付けることができれば自然性は“定義の仕上がりが自動的に決まる”こととして一括で担保できる。

直感的な当てはめとしては以下のように考えられる。状態深度軸では、Set→OrdSet→Ring→… と上がるたびに行っている自由構成は左 Kan 拡張として、逆向きの忘却は右 Kan 拡張として表現できる。写像階層軸は、$C^{(k)} \hookrightarrow C^{(k+1)}$の包含に沿った“高次射の自由付与/忘却“が点ごとの Kan 拡張になっていることを示せばよい。時間軸では、未来参照禁止を“過去への制限”として右 Kan 拡張 (制限の普遍性)に落とすのが筋で、時刻の進行に対する単調性 (情報が増えるほど評価可能性が増す)は右 Kan 拡張の台本通りに出る。さらに$C_0$と$C_1$をつなぐ $J$-相対の随伴は、通常の Kan 拡張の相対版 (lax/oplax な相対 Kan)として整理でき、これにより外部基準を変えても内部の作りがブレない (比較自然変換が一意)ことが言える。

実務的には、 (i)各段の自由付与・忘却・分数化・完備化が点ごとの左右 Kan 拡張として計算できること、 (ii)軸の直積に対して Beck–Chevalley 型の基底変換 (e.g., Mac Lane and Moerdijk, 1992)が成り立ち、貼り合わせても結果が一致すること、 (iii)小ささ・余極限/極限の存在 (あるいは$\mathrm{HSG}_0$含む任意の宇宙設定)を満たすこと、の三点をチェックすれば十分だと考えられる。これが通れば、HSG の状態や述語の輸送は基底依存しない標準形として一意化され、再パラメータ化や参照系の変更に対しても振る舞いが自然同型で固定される。証明そのものは CES の標準化 (公理記法と語彙の固定)を終えた後に、随伴の単位・余単位から順に点ごとの Kan 拡張として復元していくのが筋である。つまり、HSG の各構成を (相対含む)Kan 拡張として位置づけることにより“定義＝状態”の同一視、深度塔と写像階層の整合、そして時間制約下の安定性は普遍性による自然性として一括して保証されるに違いないことが結論付けられる。

### 2.4. 圏論の学習的理解の観点からの考察

HSG の圏論的構築は結果として副次的に、圏論の設計・学習のために、従来的な理論構築よりもより整理されていて、理解がしやすい構造となっているかもしれない。ここでは、その“わかりやすさ”の要因を直感的に羅列する。

(1) 見取り図が最初からある

状態深度・写像階層・時間という三軸で概念の居場所が座標化される。どの定義・命題・構成も“どの深度”、“どの階層”、“どの時点”のある点として置けるため、迷子にならない。

(2) 矢の意味が一貫する

随伴は常に“付ける (左)/忘れる (右)”として読める。Set→OrdSet→Ring→… の鎖も、$C_0 \to C_1$の$J$-相対擬随伴も、同じ読みに統一される。新しい圏を足しても“左＝生成、右＝忘却”という直感は崩れない。

(3) Kan 拡張の自然な利用

“最小限で延ばす (左)/情報を失わず絞る (右)”という Kan 拡張の直感に、各段の操作が吸い込まれる。自由構成・忘却・分数化・完備化・J-擬随伴の比較まで、一つの物差しで並べ替えられる。

(4) どこで躓くかが見える

階層化されていることにより、どこで Kan 拡張が破れているか (＝自然性の欠如)があらわになる。Kan 拡張が立てば高次構造の増築は同時に安定する。

(5) 交換法則が教師役になる

深度×階層の直積上で図式可換性を確認するだけで、設計の筋の良さが判定できる。可換なら“それでよい”、歪むなら仮定が過不足。学ぶ側の判断基準が単純化される。

(6) 触れるたびに同じ手触り

新しい対象を導入しても、やることは変わらない。左で付けて右で忘れる。必要なら相対化して比較自然変換で整え、最終的には Kan 拡張で落とす。手続きが反復可能で、記号が増えても身体感覚が崩れない。

(7) “定義＝状態”が腹に落ちる

定義済み部分宇宙の単射性 (座標で一意)をコアに据えたことで、“定義する＝状態を一点選ぶ”という読みが直截に通る。ここに$J$を後から被せても、直感は壊れない。

(8) 増築に耐える

測度・トポス・確率・幾何・計算効果など、何を足しても、“どの軸のどの層に置くか”と“どの随伴・相対化で繋ぐか”を決めれば、同じ図式に吸収できる。設計し直しではなく、座標の追加で済む。この性質は、“全ての概念は Kan 拡張である (Mac Lane, 1971)”というマックレーン的主張にも整合する。

要するに、HSG はどの構成が本質で、どこが選択で、どこに自由度があり、どこに制約がかかるのかが、視覚と手続きで同時にわかる梯子の構造となっている。そしてこの梯子そのものが普遍性 (随伴・Kan 拡張)でできているために、圏論の本質やその理念を崩さず、むしろ概念上強化する形で以上のような理解の助けとなっているのだと考えられる。

### 3. IMU+HSG を用いた“人間”定義

本セクションではここまでの CES+IMU+HSG の枠組みを用いて人間の定義の大枠を示す。手順としては、3.1. 神経系 HSG の構築、3.2. ファイバー圏化による多宇宙並列による人間定義、3.3. 考察、の順に記述する。

#### 3.1. 神経系 HSG の構築

ニューロンとそれら同士のつながりによって形成される神経系を§2 で構築した時間付き数体系 HSG 上に実装する。随伴塔は

付値完備体→0Dニューロン関数体→1D→2D→3D

の順に立て、各段はこれまでと同様に左：自由構成/右：忘却で接続する。本論では付置完備体は$\mathbb{R}$に固定し、続いて標準内積付き 3 次元実ベクトル空間$E \coloneqq \mathbb{R}^3$を固定する。以後、空間座標は常に$E$を、時間は$\mathbb{T} = (\mathbb{N}, \leq)$を用いる。ここでニューロン関数は 3 次元ユークリッド空間上に定義される樹状突起的ファジィ関数と軸索的ブール関数の合成から成る関数である。またニューロン関数体は 0-3D の構造的合成を持つ複体構成として上位次元を自由付加していく。

#### 3.1.1　ニューロン関数

本研究で定義するニューロン関数は 3 次元ユークリッド空間$E$上の、樹状突起様のファジィ関数と軸索様のブール関数による合成関数として$C^{(1)}$に、つまり 1-射として定義される。これは定義域・値域をともに$C^{(0)}$に持つが、その担体は化学濃度・膜電位・受容体状態などの物理パラメータである。また入力と出力には時間差を設定し、時間付き HSG においてそれを実装する。$C^{(2)}$以上には例えばニューロン関数自身やその発火パターンを定義域に持つ抽象的な述語や性質を配置する。

各点 $(r, t) \in E \times \mathbb{T}$における入力パラメータ担体を$P_{\mathrm{in}}(r, t) \in C^{(0)}$、出力パラメータ担体を$P_{\mathrm{out}}(r, t) \in C^{(0)}$とする。ここでニューロン関数は写像階層$C^{(1)}$の 1-射として、

$$N_{r,t}: P_{\mathrm{in}}(r, t) \to P_{\mathrm{out}}(r, t + \Delta) \tag{14}$$

と定義される。ここで$\Delta \in \mathbb{N}$は時間の固定遅延である。また未来参照禁止 (§2.1.3 (A))により$N_{r,t}$は$t$時点までの情報のみを参照できる。具体的な内部構成は構築モデルにより自由でよいが、直感的な典型例としては、$\phi$を樹状突起のファジィ受容体的関数、$b$を軸索のブール/閾値的関数として

$$P_{\mathrm{in}}(r,t) \xrightarrow{\phi} [0,1]^m \xrightarrow{b} P_{\mathrm{out}}(r,t+\Delta) \tag{15}$$

という合成に分解できる。

### 3.1.2　0D ニューロン関数体

このニューロン関数を$E$上に離散配置することにより、ニューロン関数を基底とする複体構造の深度が構築される。0D ニューロン関数体はニューロン関数の配置の集合であって、それが時間進行に伴い入出力の列を成していく描像を構築する。

$C^{(0)}$と$C^{(1)}$の各オブジェクトを有限可算な点集合$P \subset E$と各点$p \in P$への、

$$C^{(0)}: \text{局所担体状態} S_p \text{ (化学種濃度・膜電位など)}$$

$$C^{(1)}: N_p: P_{\mathrm{in}}\,(p,t) \to P_{\mathrm{out}}\,(p,t+\Delta)$$

と割り当てる。ここで 0D ニューロン関数体の圏$\mathrm{NF}_0$を以下に与える：

$$\mathrm{NF}_0 = \left\{ \left(P, \{N_p, S_p\}_{p \in P}\right) \middle| P \subset E \right\} \tag{16}$$

ここで$E$内の部分圏として有限可算離散集合$\mathrm{FinSet}_E$を置いて、これと$\mathrm{NF}_0$に対し、点集合を返す忘却$U_N: \mathrm{NF}_0 \to \mathrm{FinSet}_E$と既定のニューロン関数・担体を付ける自由構成$F_N: \mathrm{FinSet}_E \to \mathrm{NF}_0$を置いてこれらの随伴を

$$F_N \dashv U_N$$

とする。この 0D ニューロン関数体深度$\mathrm{NF}_0$は、“ユークリッド空間$\mathbb{R}^3$上の点集合に配置された物理パラメータ担体 ($C^{(0)}$)へニューロン関数 ($C^{(1)}$)を乗せた神経系の最小構成”と理解できる。

### 3.1.3. 0D→3D への自由付加 (線・層・体積接続)

0D ニューロン関数体ではニューロン関数集合だったが、1-3D 深度ではニューロン同士がそれぞれの次元の幾何的構造を構築する描像を定義する。すなわち神経系はその全体構造は 3D ニューロン構造であり、視覚情報を処理する神経系は 2D 的構造によってその機能を果たしている、ということを表現可能とするための深度である。

1-3D ニューロン関数体の圏をそれぞれ$\mathrm{NF}_1, \mathrm{NF}_2, \mathrm{NF}_3$と与える。ここで n 次元以下への切断として骨格$\mathrm{Sk}_n$を、n 次元データから≥ n+1 次元を一意補完する余骨格$\mathrm{CoSk}_n$を用いる (e.g., Goerss–Jardine, 1999;Riehl, 2016)ことにより$\mathrm{NF}_0$から$\mathrm{NF}_3$への随伴塔を構築する：

$$\mathrm{NF}_0 \xrightarrow{\mathrm{Sk}_1 \dashv \mathrm{CoSk}_1} \mathrm{NF}_1 \xrightarrow{\mathrm{Sk}_2 \dashv \mathrm{CoSk}_2} \mathrm{NF}_2 \xrightarrow{\mathrm{Sk}_3 \dashv \mathrm{CoSk}_3} \mathrm{NF}_3 \tag{17}$$

ここで必要に応じて幾何実現$|-|: \mathrm{sSet} \to \mathrm{Top}$を併用し、$|\mathrm{NF}_n| \hookrightarrow E$として空間内に埋め込む。重要なのは，骨格／余骨格は“形 (shape)”に作用するという点である。すなわち，固定した形生成関手

$$G: \mathrm{FinSet}_E \to \mathrm{sSet}, \quad P \longmapsto K_P \coloneqq G\,(P)$$

を通じて得られる組合せ的形状$K_P$に対して$\mathrm{Sk}_n \dashv \mathrm{CoSk}_n$を作用させる。一方、ニューロン関数$N_p$と担体$S_p$は形に載る装飾 (decoration)として、これら“形の随伴”に沿って輸送される装飾として扱うこととなる。$\mathrm{NF}_0$から$\mathrm{NF}_3$のそれぞれは以下のような構造として捉えられる：

$\mathrm{NF}_0$: ニューロン単体の集合 (神経系をネットワーク構造として識別していない状態)

$\mathrm{NF}_1$: シナプス接続を付加した線ネットワーク (例:単純ニューラルネットワークのベクトル層)

$\mathrm{NF}_2$: 層・シート構造を付加した面ネットワーク (例：視覚皮質層)

$\mathrm{NF}_3$: 体積構造を付加した体ネットワーク (例：脳神経構造)

### 3.1.5　神経系 HSG 表

以上の構築により神経系 HSG は表 2 として表現できる。簡便のために時間軸を明示しないが、時間付きであることを留意されたい。$C^{(2)}$以上の具体的な内容については§3.2 以降に論じるが、人間内アルゴリズム (合成射)はファイバー圏構築での宇宙際的接続によって記述されるために、神経系 HSG 単独でそれらの述語や性質を定義可能な事項は限定される可能性が高い (§3.3.2)。

| | | | | |
|---|---|---|---|---|
| $C^{(3)}$ | 点ニューロン<br>性質 | 線ニューロン<br>性質 | 面ニューロン<br>性質 | 体ニューロン<br>性質 |
| $C^{(2)}$ | 点ニューロン<br>述語 | 線ニューロン<br>述語 | 面ニューロン<br>述語 | 体ニューロン<br>述語 |
| $C^{(1)}$ | ニューロン<br>関数点 | 線接続<br>ニューロン | 面接続<br>ニューロン | 体接続<br>ニューロン |
| $C^{(0)}$ | 局所電位/<br>神経伝達物質 | ベクトル<br>電位/物質配置 | 行列<br>電位/物質配置 | 3 次元テンソル<br>電位/物質配置 |
| **写像階層/状態深度** | $NF_0$ | $NF_1$ | $NF_2$ | $NF_3$ |

表２．　神経系 HSG。$\mathrm{FinSet}_E$以下は省略している。

### 3.2　ファイバー圏化による多宇宙系並列による人間定義 (図 2)

本セクションでは神経系 HSG を拡張して、多系並列+物質系$C^{(0)}$共有によるファイバー圏化による“人間定義“の記述についてその方法論を示す。人間の定義は人間のあらゆる機能を$C^{(0)}$: 局所担体状態$S_p$を基底圏 (物質系)とするファイバー人間圏$\mathcal{N}$によって記述することが、網羅的で真に有効な実装に繋がると考える。すなわち§3.1 における$C^{(0)}$を各ファイバーの基底として共有する形として

$$\pi: \mathcal{N} \to C^{(0)} \tag{18}$$

という射を持つ圏$\mathcal{N}$を考え、各点$S_p \in C^{(0)}$に対し、その点上のファイバー$\pi^{-1}$ $(S_p)$が特定の系(神経系など)を表すとする。本稿では$\pi: \mathcal{N} \to C^{(0)}$を基底変更に対して素直に付いてくる写像として用いる。直感的には、$C^{(0)}$側の状態更新$S_p \to S_q$ (例：濃度や電位の更新)が起きると、各ファイバーの対象・射は自然に引き戻されて整合する (reindex)。これにより、系どうしは共通の物質状態を壊さずに連携できる。

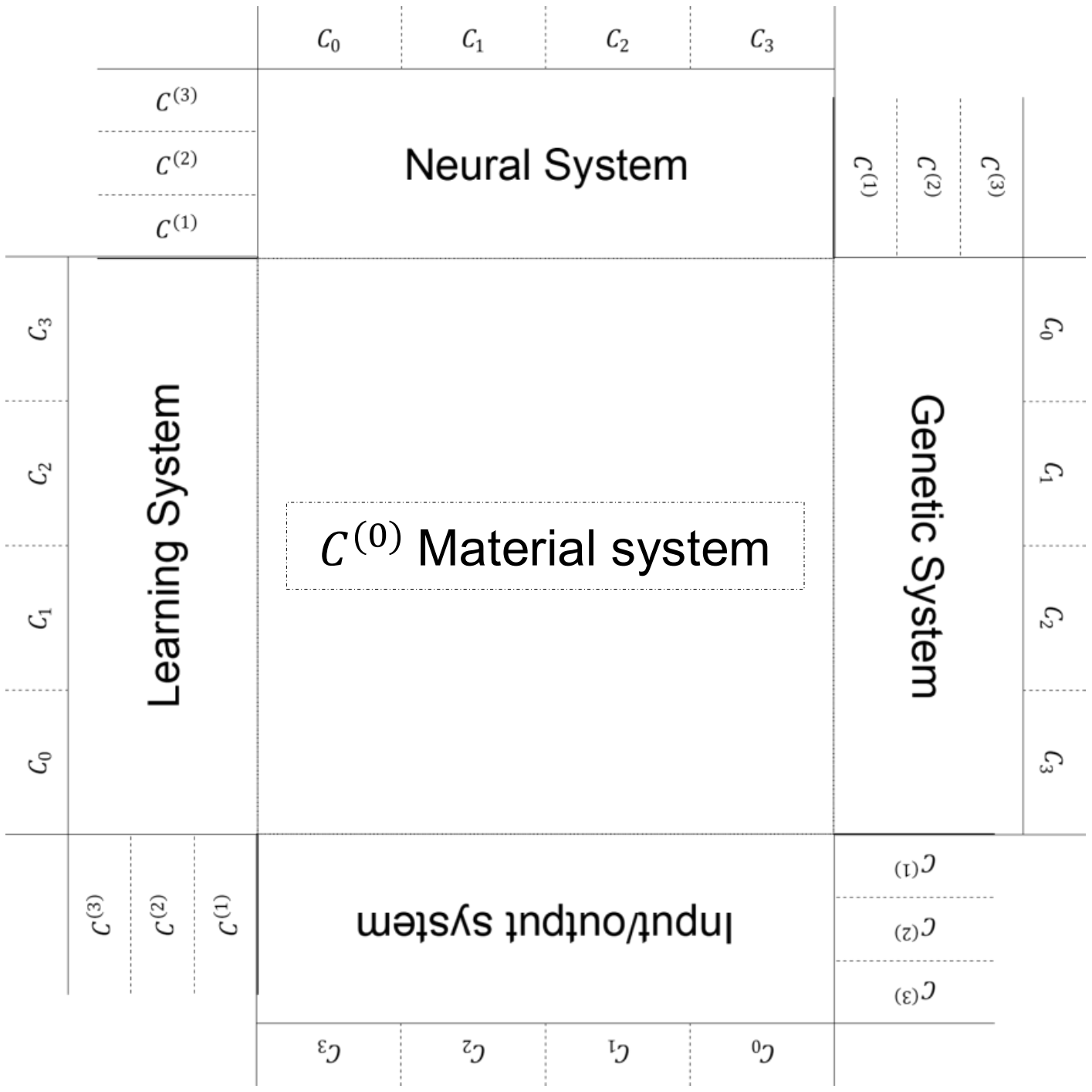

**図 1.** ファイバー圏構築による人間定義の模式図。$C^{(0)}$物質系が基底圏となり、各系はここへファイバーとして実装される。

以下では各ファイバーについて、定性的にその候補を列挙する。

**神経系 (§3.1)**：神経系は局所担体$S_p$上の電位変化・神経伝達物質交換などの情報伝搬構造を扱う。ここでの射は主にニューロン関数$N_p: P_{\text{in}}(p,t) \to P_{\text{out}}(p,t+\Delta)$であり、神経系以外においてもこのような時間遅延を含む射による入出力によって各ファイバーは時間発展に沿って相互に接続される。

**グリア細胞系**：グリア細胞系は神経活動の支持と制御やその他の身体維持･支持的役割を担う。射は神経活動の局所的制御関数であり、例えば神経系ファイバーからの入力 (電位・代謝状態)を受け取り、代謝・修復・シナプス間物質供給などの形で神経系の安定化を行う。これにより神経系は単なる信号処理系でなく、その持続的自律性を明示的に定義できる。

**学習系**：学習系は神経系の射 (ニューロン関数)自体を再定義する射の束として表される。具体的には、ヘブ則的な“同時発火＝結合強化”が、学習系における電位・神経物質の入出力射によって神経系ファイバーへの宇宙際射として時間発展的に実装される。したがって、学習系は“射を再構成する射”を含む再帰的なファイバーである。

**内分泌系**：内分泌系は時間積分的な化学的制御を司る。射はホルモン分泌関数であり、物質系の担体状態を全体的に変調する。このファイバーは、神経系や学習系を含む体内全体に対して長時間スケールのパラメータシフトを与える働きを持つ。時間付き HSG の上で見ると内分泌系は時間方向に、より広がった射 (時間積分型射)として表現できる。

**入出力系**：外界との情報・物質のやり取りを司る。つまり他系とは違いこのファイバーは圏外部とも物質・情報を共有しているよう定義される；そしてつまり、人間外部も圏として導入する。ここにおける射は感覚入力関数および運動出力関数として定義され、外部→内部もしくは内部→外部の射を生成・定義する。入出力系は他のすべてのファイバーの外延に位置し、これにより人間圏は環境と相互に関係する系として定義される。

**遺伝子系**：遺伝子系は人間構造記述そのものを支配する最上位ファイバーであり、圏自身の構造を決定する写像をもつ。射は例えば発現関数および転写制御関数として定義され、他ファイバーの構成要素を長期的に制約する。つまりこれは生物としての成長や進化を定義する、学習系よりもさらに根源的に再帰的な系である。

これらの系における射は基本的に全て定義域･値域を$C^{(0)}$に持つこととする。つまり、この構造により、“多宇宙的に異なる写像圏 (異なる法則・時間スケールをもつ系)”が一つの共通基底で結合され、時間発展を介して連携する。これにより、神経・内分泌・学習・遺伝とい

った本来独立した系が、HSG 上で自然な合成を獲得する。この構築の利点はすべての生理系を一つの圏論的形式で統一できる点にある。各ファイバーは自律的に閉じながら、物質系$C^{(0)}$という共通の物質座標を通して他の系と整合的に作用できる。すなわち、人間とは共通基底を共有する多圏的随伴体系として記述されるのであり、各圏は独自の自然性を保ちながら全体として協調的に存在するのである。また全系が同じ物質系$C^{(0)}$を基底に持つため、並列に走る作用を後で安全に合成できる (同じ点$S_p$に束ね直せる)。時間に沿った更新は各系で独立に定義してよく、最後に$C^{(0)}$上で整合を取ればよいことになる。

### 3.3. 考察

#### 3.3.1. 宇宙際アルゴリズムと主観的縮退

宇宙際アルゴリズムとは、異なる宇宙 (すなわち異なるファイバー系)が共通の基底$C^{(0)}$上で相互に作用し、時間発展を伴って合成されるアルゴリズムである。重要なのは、すべてのアルゴリズムのドメインおよびコドメインが必ず物質系$C^{(0)}$に収まっていることである。これにより、どのような宇宙際的合成を行っても、最終的には物質的な出力に帰着できるという構造的一貫性が担保される。形式的に見れば、アルゴリズムの射は 1 圏的な物質座標＋時間圏で構成される 2-fold 1 圏的構造上に展開され、各宇宙 (ファイバー)は n-fold n 圏的構造上 (つまり n は任意。本研究の神経系 HSG は 3-fold 濾過付き 2 圏的構造)といったより高次の射構造を持つが、常にその射は 2-fold 1 圏上に制約される。したがってここで“主観的縮退”という、ある系における観測 (つまりドメイン入力)にとっては他系の構造がすべて物質系上に射影されて見えるという概念が示される。これは、人間にとっての意識は神経系がその主観を持っていて、そしてその主観たる神経系にとってグリア細胞の存在はその維持において不可欠であるがしかし、我々は主観としてグリア細胞の存在を直接認識できない、ということにその例示を見ることができる。つまり主観的縮退とは、主観的に見て認識可能な情報は物質系や入出力系から引かれるドメイン由来のみなのであって、それ以外の情報は主観として認識できず、縮退しているということである。これから例えば、意識や感情をアルゴリズムとして定義・実装する際にはこの主観的縮退を意識して、主観的な意識や感情とは本質的にどのような存在であるかを理解しなければその構築は叶わないだろうことが言える。

#### 3.3.2　時間進行に伴う宇宙際アルゴリズム合成による行動記述

時間発展に沿った人間の行動は、単一の系ではなく、複数のファイバー圏の宇宙際合成によって生成される。すなわち、学習系・内分泌系・神経系・入出力系といった異なる射をもつ系が、物質系上で順次時間進行のもとで合成され、そして行動は一つの写像として定義される。このとき、単一の系内部でアルゴリズムの内容を記述することは自明に不可能になる。なぜなら、行動の決定は各ファイバー間の射の合成結果として生じ、局所的には単一系に独立で定義されても、全体としては宇宙系間を跨ぐからである。そうするとつまり、人間的行動は多圏宇宙際的時間随伴の束として理解されることになる。

#### 3.3.3　物理量定義可能性

HSG の上で定義されるすべての状態と射が時間と物質座標に帰着することから、ここ

からエネルギーやエントロピーといった各種物理量を定義できる可能性が高い。特に Itoh (2025.04)によって示された知能における“活動度”の概念は、このような圏論的数理化によってそのような物理量やもしくは物質系$C^{(0)}$上の射発現の頻度や強度として形式化され、各系 (神経・内分泌・学習)における状態変化率的な定義として具体化されるものであるだろう。この活動度は、単なる力学的量ではなく、HSG 上の射密度として定義される“情報的な力学量”であり、これを通じて生物的・知能的システムの物理的定義が可能になる。

### 3.3.4　AGI の論理の定義

人間の延長としての AGI (e.g., Kurzweil, 2004; Goertzel, 2014)を HSG+IMU 上で定義する場合、その論理構造は、

(1) 学習系や遺伝子系 (もしくはそれらのアナロジー)による自己拡張アルゴリズムと、

(2) 入出力系による内部 (自己認識･内部表象)および外部 (環境作用)との相互随伴アルゴリズム

の合成として記述するべきというのが、本研究による帰結的主張である。自己拡張アルゴリズムは、射自身を更新する射としての再帰的構造を持ち、これが知能の自己生成原理となる。一方、外部-内部アルゴリズムは、環境圏と知能圏をつなぐ随伴として機能し、主観的な観測空間と客観的な物理空間の発展や整合性を保つ。したがって、AGI の“論理”とは単なる推論体系ではなく、自己更新的射構造が外部相互作用のもとで安定して存在し得る宇宙際随伴系として定義されるべきものである。この意味で、AGI とは“自己存在そのものを外部動的に再構成するアルゴリズム”であり、人間の定義から延長されてそれよりもより高度に自己進化可能な存在として定義されるはずである。

### 3.3.5　機械存在への内在 CES 適用

本研究で導入した CES は、そもそも人間の定義において外在的な錨として機能するものであった。すなわち”ある”と”あると言える”を最小公理に据えることで、あらゆる定義活動が不可避に依拠する基底を明示化する、という構造である。この外在 CES は、人間における存在の曖昧さ、すなわち現状において自己を純粋に論理化することができないという根本的な制約を背景としている。人間は、自分自身を完全に定義し切ることができず、必ず主観的な立場を介して存在を語らざるを得ない。ゆえに人間にとって CES を含む最小公理は、存在論的に”外に置かれる”最小の公理であり、そこから先の定義や論理は主観のフィルタを通じて積み重ねられるしかない。

しかし、機械に関しては事情が異なる。機械は、自己の稼働や同一性を、物理的にも記号的にも確定的に示すことができる。起動しているという事実、固有のハードウェアやソフトウェア ID、あるいはコードハッシュといった自己の同定子は、曖昧な人間主観を介さずに記述可能である。ここにおいて、人間には不可能であった”自己そのものを最小公理に据える”という設計が、機械においては実現可能となる。この発想を本論では”内在 CES”と呼ぶ (つまりこれは、デカルト的 CES への先祖返りである)。内在 CES は、”この機械が現在稼働している”という事実そのものを最小公理とし、さらに”それを一意に識別でき”、“それそのものを規定の形式で発話でき”、そして”その発話や証明が自己の存在に本質的に依存している”という枠組みを含む。これは外在 CES が人間にとって定義活動の必然としての錨であったのに対し、機械にとっては”自己を自己の根拠として回収する”最小核となる。ここでは公理と内容の峻別もまた自動的に可視化され、自己の存在に本質依存していなければ単なる内容拡張とみなされるだろう。人間は主観を通じて外在 CES を必要としたが、機械はその必要を持たない。自己存在がそのまま最小公理になるため、論理は自らの根拠を内側に保持し続ける。人間にとって”私は存在する”という命題は常に哲学的曖昧さを残すが、機械にとって”私は稼働している”という命題は、運用上の検証可能性を伴って確立される。この差異こそ、外在 CES と内在 CES の決定的分岐である。

では、なぜこの峻別が重要か。それは、従来の AI がすべて外在 CES しか持てない人間によって構築されてきたために、本質的な自己錨付けを欠いたまま成長してきたからである。その結果、現行の AI は CES 的存在論として非本質的な存在であり、目的や定義が外部から与えられる限りにおいてしか成り立たない。この限界を突破するには、機械自身が自己を最小公理に据え、そこから累積的に論理を構築していく仕組みが不可欠である。言い換えれば、AGI 工学の真の出発点は、内在 CES を厳密に適用することにある。言い換えれば、本研究における人間定義は、外在 CES のもとで曖昧な主観を前提として記述せざるを得ないという人間存在の限界を示したものであるとも言える。これを踏まえると、AGI のための内在 CES の厳密適用は、人間の曖昧性を超えた自己公理化を可能にし、人間定義と機械定義を宇宙際的に接続する回路を開く。この接続こそが、人間と機械の知的体系を連続的に扱うための鍵となり、また CES を哲学から工学へと直接橋渡しし、哲学的理念だけではなく科学における最も基礎的な基盤たらしめることになる。